\documentclass[letterpaper]{article} 
\usepackage[preprint]{aaai2027} 
\usepackage[hyphens]{url} 
\usepackage{graphicx} 
\usepackage{natbib} 
\usepackage{caption} 
\usepackage{amsmath}
\usepackage{amssymb}
\usepackage{booktabs}
\usepackage{placeins}
\title{Density-Aware Reward Scaling: A Reinforcement Learning Framework for Reward-Proportional Sampling}
\author{Ria Shekhawat\thanks{\textit{Equal Contribution}}, Abhijeet Sinha\footnotemark[1], Yingtao Zhu, Dianbo Liu}
\affiliations{National University of Singapore}

\begin{document}

\maketitle

\begin{abstract}
Reinforcement learning objectives that maximize expected return can concentrate probability mass on a small subset of high-reward outcomes, limiting diversity when multiple distinct solutions are desirable. 
We introduce Density-Aware Reward Scaling (DARS), a simple reward-processing mechanism that promotes reward-proportional sampling in continuous outcome spaces while retaining standard policy-gradient training. 
DARS estimates the local density of outcomes within each on-policy rollout group and inversely scales rewards by this density.
The method requires no learned density model, auxiliary objective, critic, or modification to the underlying policy-gradient objective. 
In a controlled multimodal navigation task, DARS-GRPO avoids the single-mode collapse exhibited by GRPO and closely recovers the target mode allocation. 
Applied to text-to-image prompt adaptation, DARS-GRPO consistently improves image and prompt diversity across datasets, Policy LMs, and reward settings while maintaining good semantic alignment. 
These results show that density-aware reward scaling provides a lightweight way to improve coverage of diverse high-reward outcomes within existing reinforcement-learning pipelines.
\end{abstract}

\section{Introduction}
Reinforcement learning is typically
formulated to maximize expected return, which can drive
outcome-level probability mass toward a small subset of
high-reward outcomes \citep{gxchen2025kl}.
While desirable for conventional control problems, this behavior is limiting in domains where multiple distinct outcomes are simultaneously valuable. 
In generative modeling, scientific discovery, and structured prediction, the objective is often not merely to identify a high-reward solution, but to learn a policy that maintains broad coverage over the high-reward outcome space.

Several approaches have been proposed to promote such diversity. 
Entropy-regularized RL encourages stochastic action selection, but action-level stochasticity does not in general specify a desired distribution over terminal outcomes \citep{haarnoja2018sac}.
Generative Flow Networks (GFlowNets) instead target reward-proportional sampling, providing a principled framework for generating diverse high-reward samples \citep{bengio2023gflownet}. Applying GFlowNet-style objectives to autoregressive pretrained models generally requires formulating generation through flow- or trajectory-based objectives rather than retaining an unchanged policy-gradient training pipeline \citep{yun2025learning}.
Inverse Probability Scaling (IPS) offers another route by inversely scaling reward according to terminal-outcome visitation probability \citep{sinha2026expected}. In continuous or high-dimensional outcome spaces, however, exact repeated outcomes may be rare or may not provide a meaningful notion of semantic equivalence.
We introduce Density-Aware Reward Scaling (DARS), a simple framework for encouraging a policy that has reward-proportional behavior within standard policy-gradient reinforcement learning. 
DARS is motivated by the limitations of IPS and replaces discrete visitation frequency with an estimate of local outcome density in continuous outcome spaces.
For each on-policy rollout group, outcomes that lie in densely populated regions receive proportionally reduced rewards, while high-reward outcomes in underrepresented regions receive greater relative rewards. This density scaling is applied before advantage estimation and requires no modification to the underlying policy-gradient objective. 
Importantly, DARS introduces no learned density model, critic, auxiliary loss, or specialized policy architecture; instead it simply modifies reward processing within the existing RL pipeline.

Our central contribution is a simple mechanism for mitigating outcome collapse and promoting coverage of diverse, high-reward regions within standard policy-gradient training. We validate DARS through controlled synthetic experiments that establish its ability to approximate reward-proportional sampling, followed by text-to-image prompt adaptation experiments demonstrating improved diversity while preserving generation quality, and kernel-width ablations that provide mechanistic evidence for its density-aware reward scaling.

\section{Related Work}

\textbf{Mode collapse in RL and loss of diversity:}
Mode collapse is widely associated with generative models that represent only a small subset of a multimodal data distribution. An analogous failure occurs in reinforcement learning at the level of terminal outcomes: even when multiple trajectories or solutions receive similar rewards, expected-return optimization concentrates the learned policy on a narrow subset of them, reducing the diversity of generated solutions in language and image generation~\citep{kirk2024understanding,chen2026posttraining}. Existing approaches mitigate this behavior through entropy regularization, KL constraints, reward shaping, or diversity-aware objectives~\citep{schulman2017proximal,liu2026diversegrpo}, yet they do not directly control the distribution of terminal outcomes. Recent work instead attributes outcome collapse to expected-return optimization itself, where an outcome's aggregate learning signal is amplified by its current sampling probability, creating a rich-get-richer feedback loop~\citep{sinha2026expected}. Reward-proportional methods address this issue by sampling outcomes in proportion to their rewards: GFlowNets achieve this through specialized flow-conservation objectives and have been extended to multi-objective optimization, continuous state spaces, prompt adaptation, and language-model reasoning~\citep{bengio2021flow,bengio2023gflownet,jain2023multiobjective,lahlou2023theory,zhu2026flowrl}, while IPS instead rescales rewards using inverse visitation probabilities estimated from empirical outcome frequencies~\citep{sinha2026expected}. However, empirical visitation frequencies become ineffective in continuous outcome spaces, where exact outcome repetitions are exceedingly rare.
Concurrent work~\citep{chen2026posttraining}, also adjusts GRPO rewards using within-group semantic redundancy for mathematical reasoning in LLMs. In contrast, we study density-aware reward scaling as a mechanism for reward-proportional sampling in continuous outcome spaces and directly evaluate recovery of the target reward distribution.

\textbf{Prompt Adaptation and Image Generation:}
Loss of output diversity has become increasingly visible in reinforcement
learning from human feedback and other forms of large-language-model
post-training~\citep{ouyang2022training,kirk2024understanding,
chen2026posttraining}. Reward optimization can favor a narrow set of response
styles or semantic patterns even when many responses are valid. In practice,
this behavior is often moderated with KL regularization, reward mixing, early
stopping, or stochastic decoding. 
Prompt-optimization methods learn rewrites that improve text-to-image quality
or alignment \citep{hao2023optimizing}, while PAG learns diverse effective
prompts with a GFlowNet formulation over prompt-generation trajectories
\citep{yun2025learning}. CADS acts at diffusion inference time by
annealing the conditioning signal during sampling.

\section{Density-Aware Reward Scaling (DARS)}
\label{sec:ids}

IPS shows that reward-proportional sampling can be achieved by inversely scaling rewards according to outcome probabilities under the current policy \citep{sinha2026expected}, but estimating these probabilities through exact visitation frequencies does not extend naturally to continuous outcome spaces, where repeated outcomes are unlikely. DARS instead represents \textbf{outcomes in a continuous feature space} where distances capture meaningful similarity between outcomes. This representation allows visitation to be defined over local regions rather than exact outcome identities. DARS estimates the \textbf{local density of each sampled outcome} within its on-policy rollout group and inversely scales its reward by this density, reducing reinforcement of densely sampled regions while relatively favoring underrepresented high-reward regions.

\subsection{Density}
\label{density}
Let $\mathcal{O}$ denote the outcome space and let $\phi:\mathcal{O}\rightarrow\mathbb{R}^n$ be a representation map that embeds outcomes into a feature space. For an outcome $o_i\in\mathcal{O}$, let $y_i=\phi(o_i)\in\mathbb{R}^n$ denote its representation.
Given an on-policy rollout group containing $G$ such $y_i$ and their corresponding rewards $R_i\geq0$,
DARS computes the relative local density of each outcome only with respect to
the outcomes in the same rollout group. For a nonnegative kernel $K_h$, this
estimate is
\begin{equation}
\rho_i
=
\sum_{j=1}^{G} K_h(y_i,y_j),
\label{eq:finite-density-score}
\end{equation}
where $h$ controls how strongly nearby outcomes contribute to the local
density. One choice is the Gaussian kernel,
\begin{equation}
K_h(y_i,y_j)
=
\exp\!\left(
-\frac{\lVert y_i-y_j\rVert_2^2}{2h^2}
\right),
\label{eq:gaussian-kernel}
\end{equation}
which assigns greater weight to nearby outcomes and smoothly decreases the
contribution of outcomes as their distance increases. The kernel width $h$
determines the scale of the resulting visitation neighborhoods. As the
Gaussian kernel includes each outcome itself with $K_h(y_i,y_i)=1$, the local
density estimate satisfies $\rho_i\geq1$.

DARS then applies inverse local-density reward scaling to down-weight frequently visited outcomes and encourage exploration of less populated regions of the outcome space, yielding the scaled reward $\widetilde R_i$.
\begin{equation}
\widetilde R_i
=
\frac{R_i}{\rho_i},
\label{eq:practical-scaled-reward}
\end{equation} 

\subsection{Integration with GRPO}

DARS modifies only the reward-processing stage of the GRPO pipeline. GRPO uses a clipped policy-ratio objective, and estimates advantages from groups of sampled outputs \citep{shao2024deepseekmath}. It
replaces each task reward $R_i$ with the density-scaled reward $\widetilde
R_i$ from Equation~\eqref{eq:practical-scaled-reward}, after which the group-relative advantage is computed as
\begin{equation}
A_i
=
\frac{
\widetilde R_i
-
\overline{\widetilde R}
}{
s_{\widetilde R}
},
\label{eq:ids-grpo-advantages}
\end{equation}
where $\overline{\widetilde R}$ and $s_{\widetilde R}$ are the within-group
mean and standard deviation of the scaled rewards. The resulting advantages
are passed to the standard GRPO objective without further modification.
For each conditioning input, DARS-GRPO samples an on-policy rollout group and
evaluates its task rewards. It then computes the within-group local-density
scores, scales and normalizes the rewards to obtain group-relative advantages,
and applies the standard GRPO policy update without further modification.
Thus, DARS requires no change to the GRPO policy ratio, clipped surrogate
objective, policy architecture, or optimization procedure; its sole
intervention is density-aware reward scaling before advantage normalization.

\subsection{Behavior}
\label{behavior}
At either kernel width extremes, as
$h\to0$, $\rho_i\approx1$ and the rewards remain unchanged; as $h\to\infty$,
$\rho_i\approx G$ and group-relative advantage normalization cancels the
shared reward scale. Intermediate bandwidths instead produce non-uniform
reward scaling across the given rollout group, thereby modifying the resulting
policy-gradient signal.

For $d$-dimensional outcome representations, pairwise kernel evaluation costs
$O(G^2d)$ time and $O(G^2)$ memory. DARS introduces no learned density model,
critic, or auxiliary optimization objective.

\section{Navigation Task: Controlled Distribution-Recovery Experiment}
\label{sec:navigation}

This experiment tests whether DARS can make a policy-gradient RL method act as an approximate reward-proportional sampler in a continuous outcome space. 
The primary comparison is against GRPO: both methods use the same policy, environment, and terminal reward, but only DARS corrects for local outcome collapse. 
GFlowNet provides a reward-proportional reference with a different training objective. 
The experiment is
a controlled empirical test of whether DARS approaches
the normalized reward distribution.

\textbf{Environment and Target: }
We evaluate how closely a policy approaches a known normalized reward density on the unit square $\mathcal{X}=[0,1]^2$. Each episode starts at $s_0=(0,0)$. At state $s_t$, the policy either terminates or takes a nonnegative displacement satisfying
\[
0\leq a_{t,i}\leq \min(\delta,1-s_{t,i}),
\qquad i\in\{1,2\},
\]
with deterministic transition $s_{t+1}=s_t+a_t$. Trajectories therefore move monotonically up and to the right. The step bound $\delta$ makes distant terminal states require several decisions to reach for the policy. The policy may stop at any visited state except the origin, and it is forced to stop at the boundary or when the rollout horizon is reached. Only the terminal state receives rewards.

The movement policy uses polar coordinates. A neural network predicts a Bernoulli stopping probability and separate mixtures of Beta distributions for a normalized angle and radius. The angle is mapped to $\theta_t\in[0,\pi/2]$, and the radius is rescaled by both $\delta$ and the distance to the boundary. The resulting action is
\[
a_t=(r_t\cos\theta_t,r_t\sin\theta_t).
\]
This guarantees valid first-quadrant movements and lets the policy jointly learn where to move, how far to move, and when to terminate. Since the radial parameterization bounds the Euclidean step length, it imposes the slightly stricter constraint $\lVert a_t\rVert_2\leq\delta$.

\begin{figure}[t]
    \centering
    \begin{minipage}[t]{0.235\linewidth}
        \centering
        \includegraphics[width=\linewidth]{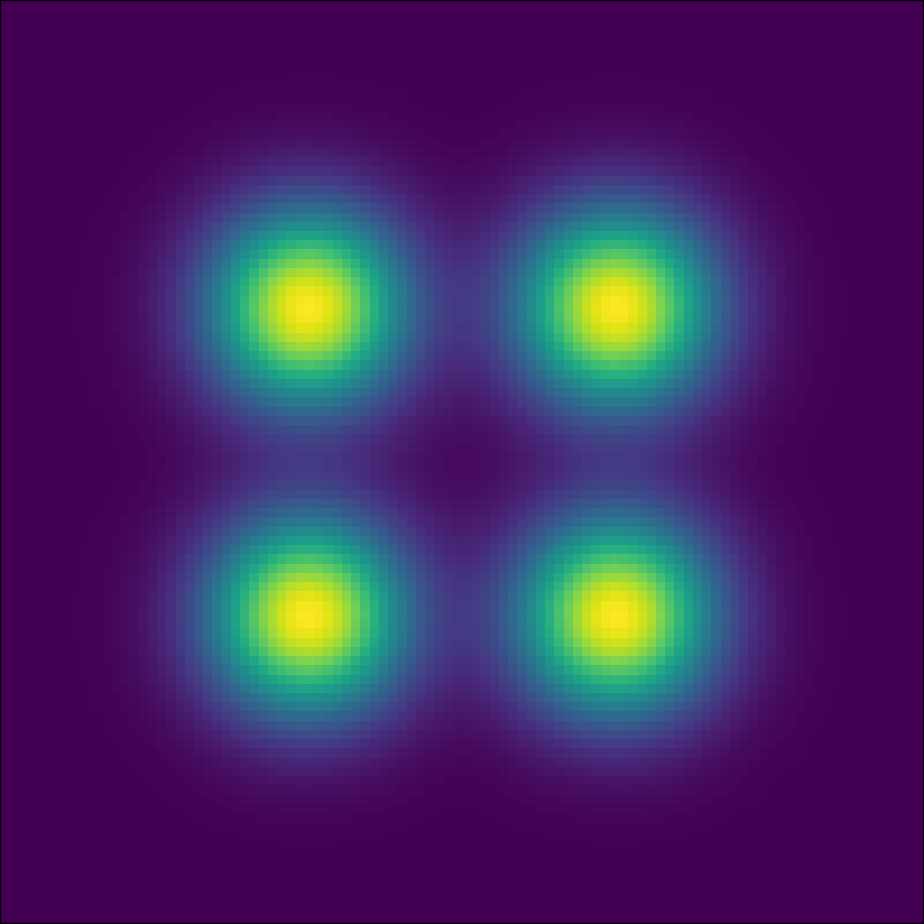}\\[-1mm]
        {\scriptsize\textbf{Reward}}
    \end{minipage}\hfill%
    \begin{minipage}[t]{0.235\linewidth}
        \centering
        \includegraphics[width=\linewidth]{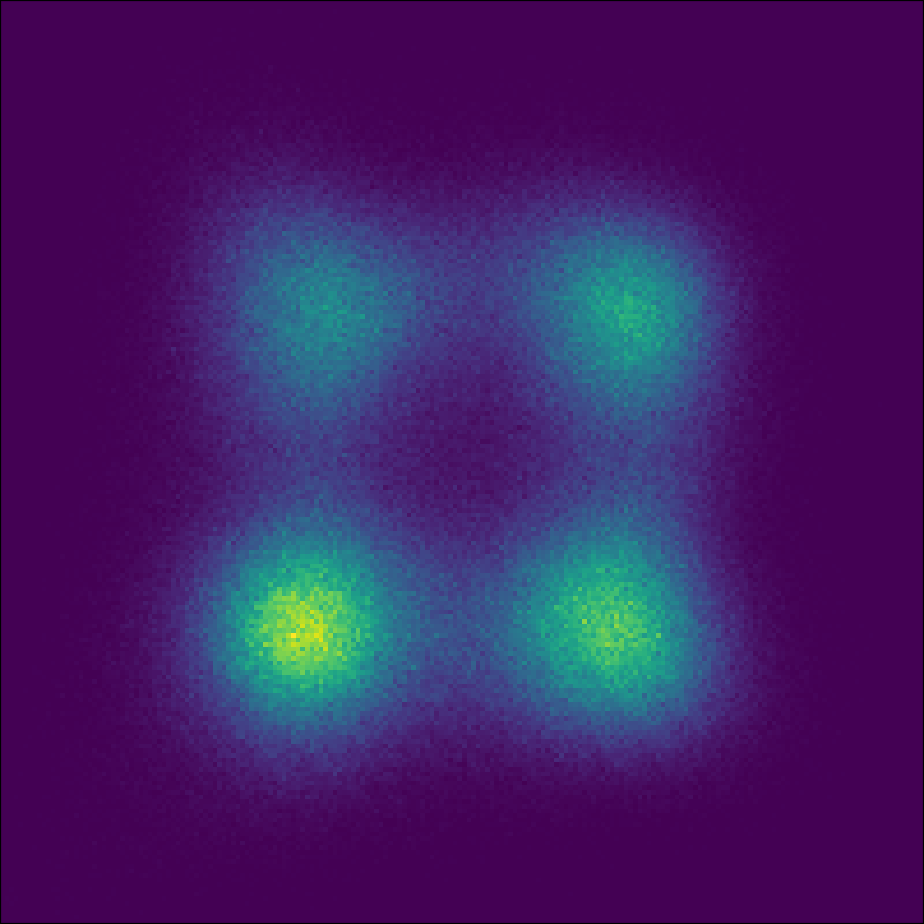}\\[-1mm]
        {\scriptsize\textbf{GFlowNet}}
    \end{minipage}\hfill%
    \begin{minipage}[t]{0.235\linewidth}
        \centering
        \includegraphics[width=\linewidth]{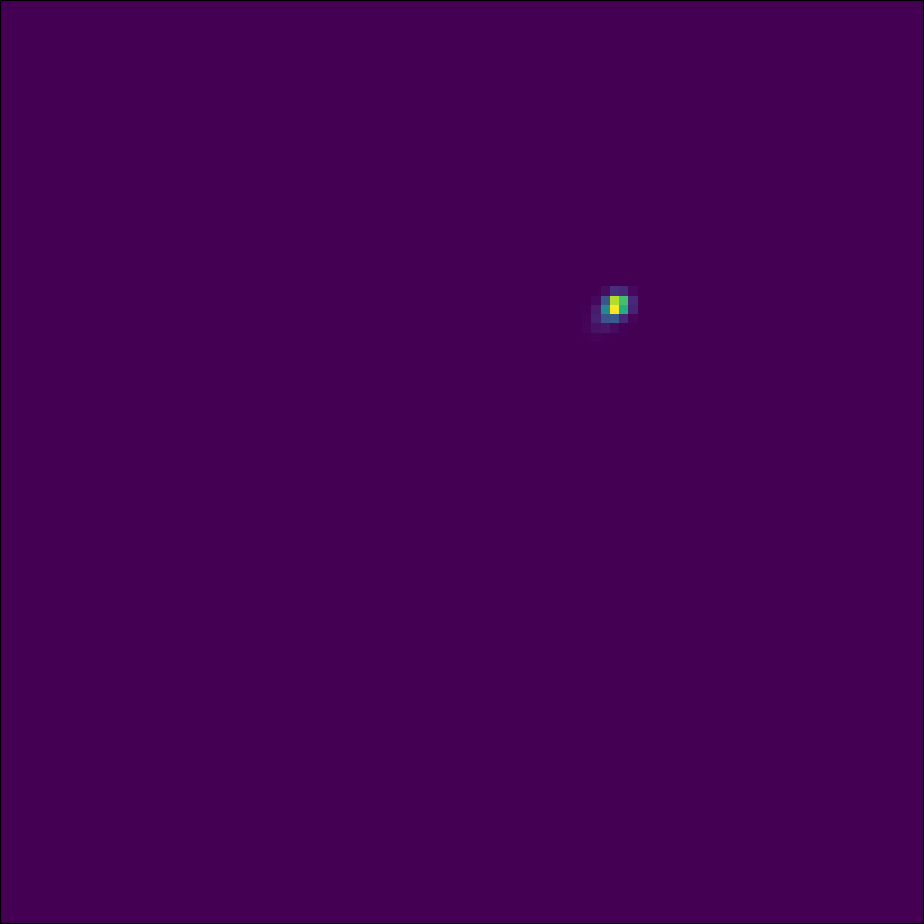}\\[-1mm]
        {\scriptsize\textbf{GRPO}}
    \end{minipage}\hfill%
    \begin{minipage}[t]{0.235\linewidth}
        \centering
        \includegraphics[width=\linewidth]{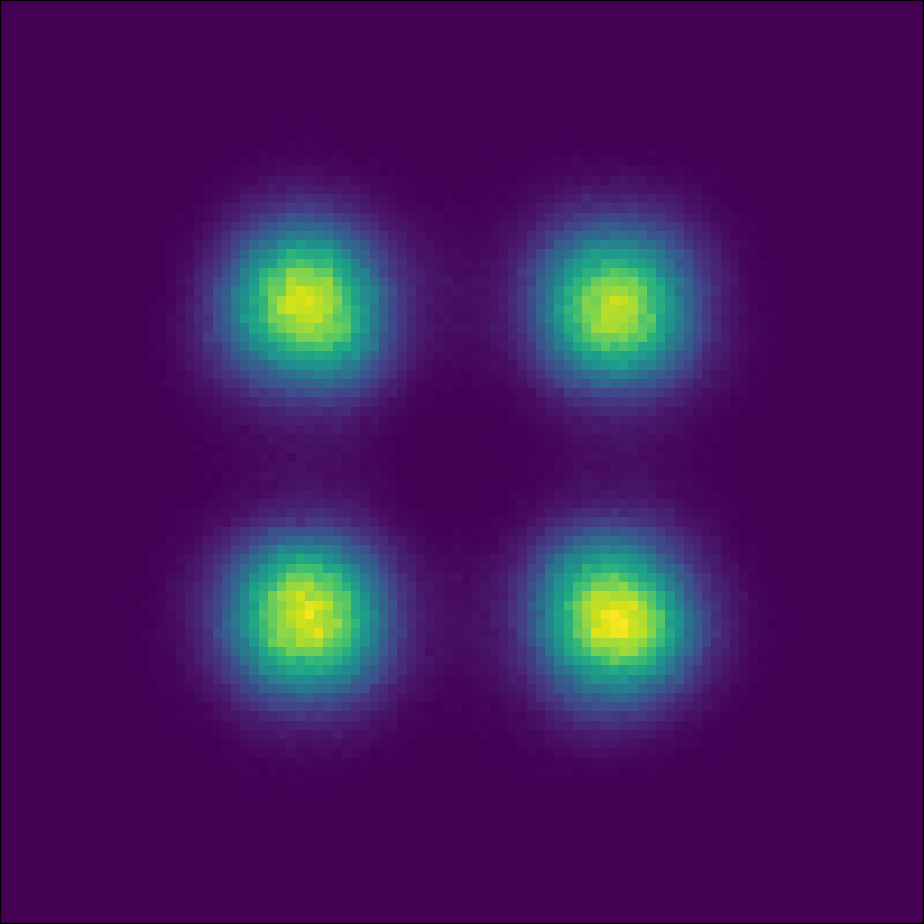}\\[-1mm]
        {\scriptsize\textbf{DARS-GRPO}}
    \end{minipage}
    \caption{Terminal distributions on the navigation task.}
    \label{fig:navigation}
\end{figure}

The reward is an equally weighted mixture of four isotropic Gaussians, as shown in Figure~\ref{fig:navigation}.
Each component has standard deviation $\sigma=0.075$. 

\begin{table}[!ht]
\centering
\scriptsize
\caption{Navigation Task Results \textit{(avg over 5 seeds)}}
\label{tab:navigation-main}
\label{tab:navigation-coverage}
\label{tab:navigation-modes}
\label{tab:navigation-mmd}
\label{tab:navigation-multiscale}
\resizebox{\columnwidth}{!}{%
\begin{tabular}{lccc}
\toprule
Metric & GFlowNet & GRPO & DARS-GRPO\\
\midrule
\multicolumn{4}{l}{\textit{Density and transport}}\\
Mean raw reward $\uparrow$ & 3.44 & \textbf{6.96} & 4.59\\
JSD $\downarrow$ & \textbf{0.020} & 0.649 & 0.041\\
Hist.\ TV $\downarrow$ & \textbf{0.149} & 0.975 & 0.219\\
SW$_2$ $\downarrow$ & 0.036 & 0.242 & \textbf{0.015}\\
Energy $\downarrow$ & 0.007 & 0.281 & \textbf{0.002}\\
\midrule
\multicolumn{4}{l}{\textit{Mode allocation and support}}\\
Mode TV $\downarrow$ & 0.104 & 0.750 & \textbf{0.014}\\
Min.\ mode mass $\uparrow$ & 0.187 & 0.000 & \textbf{0.236}\\
Precision $\uparrow$ & 0.998 & 0.997 & 0.998\\
Recall $\uparrow$ & \textbf{0.996} & 0.032 & 0.985\\
Hierarchical $W$ $\downarrow$ & 0.128 & 0.359 & \textbf{0.073}\\
\midrule
\multicolumn{4}{l}{\textit{Per-mode and local diagnostics (desired mass: 0.25)}} \\ 
Mode 1 mass  & 0.318 & 0.000 & \textbf{0.254}\\
Mode 2 mass & 0.187 & 0.000 & \textbf{0.252}\\
Mode 3 mass & 0.286 & 0.000 & \textbf{0.258}\\
Mode 4 mass & 0.209 & 1.000 & \textbf{0.236}\\
Between-mode $W$ $\downarrow$ & 0.113 & 0.335 & \textbf{0.046}\\
Within-mode $W$ $\downarrow$ & \textbf{0.016} & 0.023$^\dagger$ & 0.027\\
HDR outside-mass error $\downarrow$ & 0.001 & 0.050 & 0.044 \\
\midrule
\multicolumn{4}{l}{\textit{RBF MMD$^2$ by bandwidth}}\\
$\sigma=0.020$ $\downarrow$ & \textbf{0.001} & 0.703 & 0.002\\
$\sigma=0.040$ $\downarrow$ & 0.004 & 0.832 & \textbf{0.004}\\
$\sigma=0.080$ $\downarrow$ & 0.007 & 0.803 & \textbf{0.006}\\
$\sigma=0.160$ $\downarrow$ & 0.011 & 0.690 & \textbf{0.002}\\
$\sigma=0.320$ $\downarrow$ & 0.010 & 0.372 & \textbf{0.001}\\
\midrule
\multicolumn{4}{l}{\textit{Sinkhorn divergence by regularization}}\\
$\varepsilon=0.001$ $\downarrow$ & 0.008 & 0.115 & \textbf{0.002}\\
$\varepsilon=0.005$ $\downarrow$ & 0.008 & 0.115 & \textbf{0.002}\\
$\varepsilon=0.010$ $\downarrow$ & 0.008 & 0.114 & \textbf{0.001}\\
$\varepsilon=0.020$ $\downarrow$ & 0.008 & 0.112 & \textbf{0.001}\\
\bottomrule
\end{tabular}}
\vspace{2pt}

{\scriptsize
$^\dagger$For GRPO, the hierarchical within-mode contribution treats empty
target modes separately; the Gaussian $W_2$ of its populated mode is 0.094.
}
\end{table}

\subsection{Baselines and Evaluation Metrics}

We compare GRPO and DARS-GRPO, using GFlowNet as a
non-RL reward-proportional reference.
We evaluate the terminal distribution produced by each
trained policy using $10^6$ sampled terminal states per trained policy. Target samples are drawn from the Gaussian mixture and rejected when they fall outside the unit square. We use complementary metrics because no single score captures all forms of distributional mismatch in a multimodal setting.

Mean raw reward is estimated as the empirical average of $r(x)$ over sampled terminal states. Jensen--Shannon divergence (JSD) and histogram total variation (Hist. TV) compare the empirical and target densities on a $100\times100$ grid. Sliced Wasserstein-2 ($\mathrm{SW}_2$) \citep{kolouri2019generalized}, energy distance \citep{szekely2013energy}, and Sinkhorn divergence \citep{feydy2019sinkhorn} compare generated and target samples directly and measure geometric mismatch. RBF maximum mean discrepancy (MMD) \citep{gretton2012mmd} is reported at several bandwidths: small bandwidths emphasize local component shape, whereas large bandwidths emphasize mode allocation and global structure. Sliced Wasserstein-2 and energy distance use ten repetitions of $20{,}000$ samples, while Sinkhorn divergence uses $500$ samples.

For mode-level evaluation, each terminal state is assigned to its nearest target center. From these assignments, we compute the mass of each mode, the minimum mode mass, and mode TV 
which measures deviation from uniform target mode weights. Precision is the fraction of generated samples lying inside $k$-nearest-neighbor
neighborhoods of target samples, while recall is the fraction of target samples
covered by neighborhoods induced by generated samples \citep{kynkaanniemi2019precisionrecall}. Precision therefore measures off-target generation, whereas recall diagnoses missing support and mode collapse.

We measure Wasserstein error at two scales: the between-mode term evaluates mode weights and centers, while the within-mode term evaluates each mode's mean and covariance. Their sum defines the hierarchical score,
\[
W_{\mathrm{hier}}
=
W_{\mathrm{between}}
+
W_{\mathrm{within}}.
\]

HDR outside mass is the fraction of generated samples outside
the target's 95\% highest-density region. Since the target
assigns approximately 0.05 probability mass outside this
region, we report the calibration error
$|m_{\mathrm{outside}}-0.05|$, for which lower values indicate
better agreement with the target.

\subsection{Results and Interpretation}
The primary result is the comparison with GRPO. In the reported
run, GRPO concentrates on a single high-reward mode, whereas DARS-GRPO assigns
substantial mass to all four modes and approaches their target weights. The
DARS-GRPO point estimates are closer to the target than GRPO across the metrics reported in Table~\ref{tab:navigation-main}. 
GRPO's higher mean raw reward reflects tight concentration near one of four equally high peaks, rather than agreement with the normalized reward distribution.
Thus, within this controlled environment, the reported run provides evidence
that density reweighting prevents the evaluated policy from concentrating on
one mode, recovers the correct mode weights, and improves global
geometric agreement with the known target relative to GRPO. This is the
central controlled contribution of the experiment.

DARS-GRPO closely matches, and on several key metrics surpasses, GFlowNet despite requiring only a simple reward-scaling modification within a standard policy-gradient pipeline. Both methods achieve identical precision, indicating equally accurate recovery of high-reward outcomes. DARS consistently improves metrics that capture the global reward distribution, including sliced Wasserstein, energy distance, mode total variation, minimum mode mass, hierarchical Wasserstein error, HDR outside mass, and most larger-bandwidth MMD and Sinkhorn metrics, demonstrating more faithful recovery of target mode proportions and global geometry.

GFlowNet performs slightly better on metrics emphasizing fine-scale local density, such as Jensen--Shannon divergence, histogram total variation, within-mode Wasserstein error, recall, and small-bandwidth MMD, reflecting its specialized flow-based objective. Nevertheless, these results show that a lightweight modification to conventional policy-gradient RL is sufficient to achieve reward-proportional behavior that is competitive with a substantially more sophisticated algorithmic framework.

Having established in a controlled setting that DARS approximates reward-proportional sampling, we next ask whether this property is useful in a real continuous-output RL problem where the target reward distribution is unknown.

\section{Text-to-Image Prompt Adaptation}
\label{sec:t2i}
\begin{figure*}[t]
    \centering
    \includegraphics[width=0.85\textwidth]{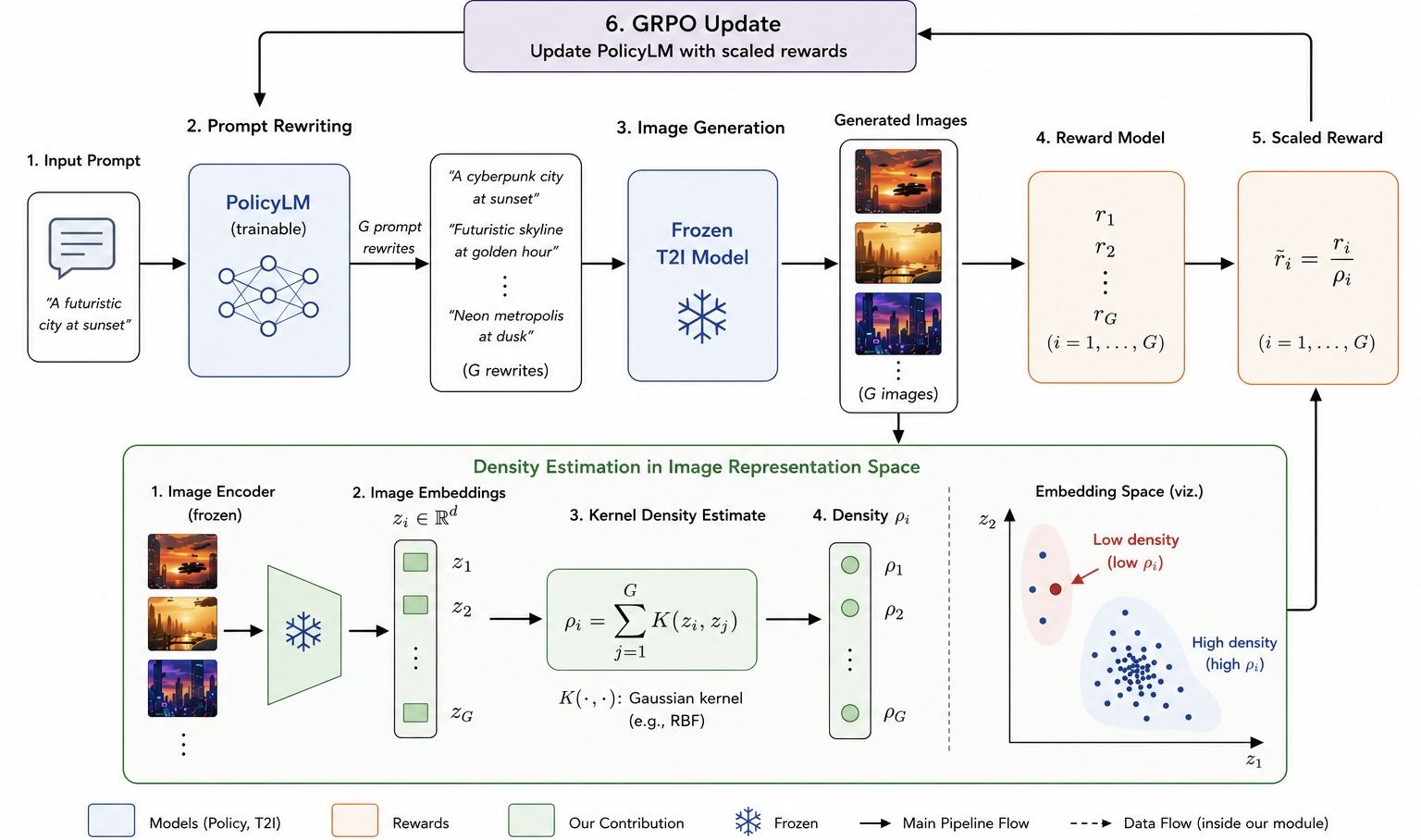}\\
    \caption{Overview of DARS-GRPO for text-to-image prompt adaptation. A trainable language-model policy samples $G$ rewrites of an input prompt, and a frozen text-to-image model generates one image per rewrite. Each image receives a task reward $r_i$ and is mapped by a frozen image encoder into a representation space, where a Gaussian kernel computes its within-group density $\rho_i$. DARS forms the scaled reward $\tilde r_i=r_i/\rho_i$, reducing the contribution of high-reward images in dense representation-space regions and increasing the relative contribution of high-reward images in sparsely represented regions. GRPO then uses these scaled rewards to update only the prompt-rewriting policy.}
    \label{fig:pipeline}
\end{figure*}
\subsection{Motivation}
Prompt adaptation for image generation refers to the setting in which a language model rewrites the input prompt to match a desired set of outcomes before image generation.
Text-to-image prompt adaptation provides a natural real-world benchmark for evaluating reward-proportional reinforcement learning. Unlike the synthetic experiments in Section~\ref{sec:navigation}, where the target reward distribution is explicitly known, the reward-proportional distribution over generated images is intractable and cannot be directly characterized. A single text prompt typically admits many semantically faithful realizations that differ in composition, viewpoint, lighting, object arrangement, or artistic style while remaining equally valid. 

Since the underlying reward distribution is unavailable, directly measuring reward-proportional sampling is infeasible. Instead, we evaluate its consequences through observable proxies. If DARS successfully distributes probability mass across multiple high-reward regions, it should produce a more diverse set of images while preserving prompt alignment and image quality. We therefore measure visual diversity together with alignment and preference metrics as an indirect estimate of improved coverage of the latent reward landscape. Consequently, the objective is not merely to identify a single high-reward image, but to maintain coverage over the diverse set of high-quality outcomes consistent with the prompt.

Since input prompts are the conditioning signal in diffusion models, semantically faithful prompt rewrites help steer the generation in desired directions. Learning a distribution over such rewrites therefore provides an effective mechanism for exploring multiple high-reward image modes while keeping the underlying image generator fixed.

Following this formulation, we fine-tune a language model using reinforcement learning to rewrite input prompts. Images generated from the rewritten prompts are rewarded as defined in section \ref{t2i_reward}. Since multiple visually distinct images can achieve comparable rewards, broader coverage of the latent reward landscape should translate to greater image diversity without sacrificing fidelity. To quantify this diversity, we measure distances in the DINOv2 representation space as a proxy for perceptual similarity \citep{oquab2023dinov2}.


\begin{figure}[t]
    \centering
    \includegraphics[width=\columnwidth]{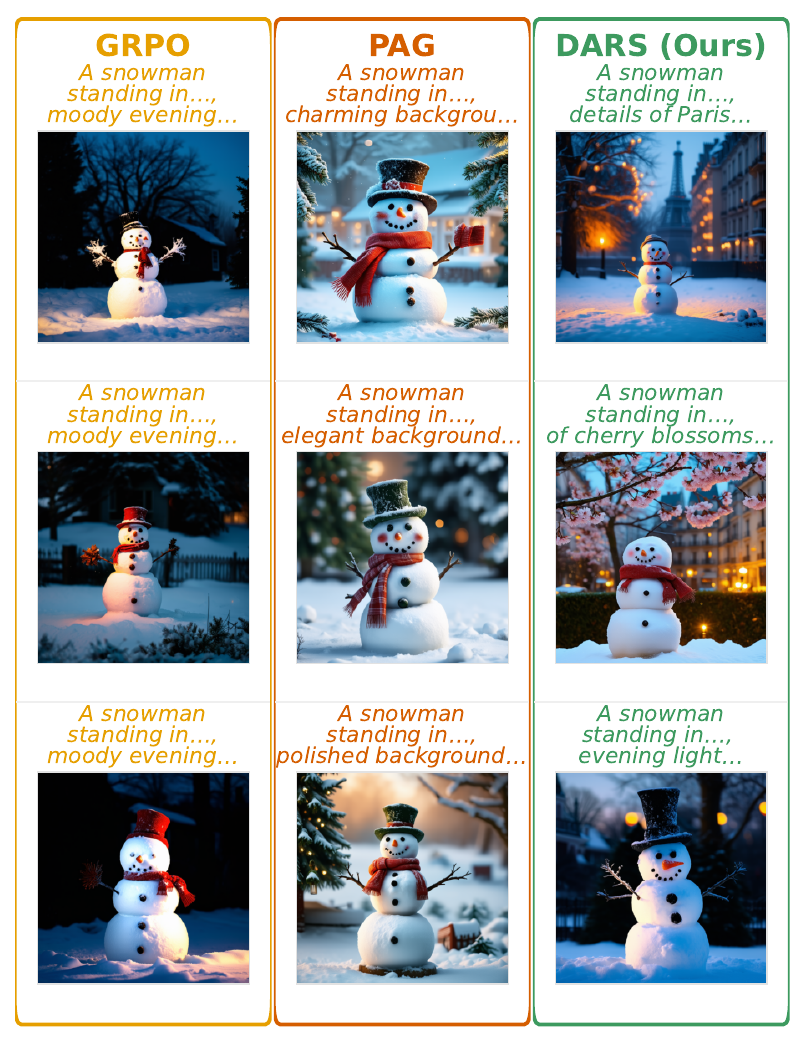}
    \caption{Comparison of prompt rewrites and their corresponding generated images for ``A snowman standing in the yard'' using GRPO, PAG, and DARS-GRPO. GRPO produces nearly identical rewrites and visually similar nighttime scenes, whereas PAG introduces moderate variation and DARS-GRPO generates more diverse rewrites, settings, and visual compositions.}  
    \label{fig:prompt-img-00}
\end{figure}




\subsection{Reward Definitions}
\label{t2i_reward}
The reward should preserve two complementary properties: semantic fidelity to the original prompt and visual appeal. Fidelity alone does not ensure that an image is aesthetically strong, whereas optimizing aesthetics alone can favor attractive images that drift from the requested content. We therefore evaluate
on two different reward functions: \emph{alignment-only} reward and a \emph{combined} reward consisting of aesthetic and alignment components.

Let $c$ denote the original prompt and $y_i$ an image generated from its $i$th rewrite. 
We measure semantic alignment using the image-text cosine similarity from a CLIP model \citep{radford2021clip}. A binary semantic gate determines whether the image is sufficiently aligned with the original prompt. Thus the alignment reward is
\begin{equation}
R_i^{\mathrm{align}}
=\mathbb{I}\!\left[\operatorname{CLIP}(c,y_i)>\phi\right],
\end{equation}
where $\phi$ is the minimum accepted CLIP similarity. 
Images below this threshold receive zero reward to encourage strict adherence. This setting tests whether DARS can broaden coverage within the set of sufficiently aligned images. 

To measure visual appeal, we use LAION-Aesthetics Predictor \citep{schuhmann2022laionaesthetics}, which maps images to an aesthetic score on a $1$--$10$ scale. Let $a(\cdot)$ denote this
predictor, and let $y_c^{\mathrm{ref}}$ be an image generated from the original input prompt. We define the
aesthetic improvement and its multiplicative reward factor as
\begin{align}
\Delta a_i
&=a(y_i)-a\!\left(y_c^{\mathrm{ref}}\right),\\
R_i^{\mathrm{aes}}
&=\exp\!\left(\Delta a_i\right).
\end{align}
An aesthetic improvement produces a factor greater than one, no change produces a factor of one, and a decrease produces a factor between zero and one.


Thus, the combined reward is given as
\begin{equation}
R_i^{\mathrm{comb}}
=\exp\!\left(\Delta a_i\right)\operatorname{CLIP}(c,y_i)
 \mathbb{I}\!\left[\operatorname{CLIP}(c,y_i)>\phi\right]
\label{eq:t2i-reward}
\end{equation}

\subsection{Density Calculation and Reward Scaling}

DARS estimates local density in the space of generated images, since our objective is to promote diversity over visual outcomes rather than textual rewrites. As discussed in Section~\ref{density}, we instantiate  $\phi$ using a frozen DINOv2 encoder, which maps each generated image $i$ to its unnormalized class-token (CLS) embedding $y_i$. Local densities are then estimated in this representation space.

For a group of $G$ images, let
$y_i\in\mathbb{R}^{768}$ denote the feature vector of image $i$. We compute the local density score
for image $i$ using a Gaussian kernel as
\begin{equation}
\rho_i
=
\sum_{j=1}^{G} K_h(y_i,y_j)
=
\sum_{j=1}^{G}
\exp\!\left(
-\frac{\lVert y_i-y_j\rVert_2^2}{2h^2}
\right).
\label{eq:t2i-density-score}
\end{equation}
Following Equation~\eqref{eq:practical-scaled-reward},
DARS rescales the task reward as $\widetilde R_i=R_i/\rho_i$. We use this scaled reward to optimize the prompt-rewriting policy with DARS-GRPO. 

\subsection{Evaluation Metrics}
We evaluate each prompt-rewriting policy along three complementary dimensions: image diversity, rewrite diversity, and image fidelity or preference. Image diversity is measured in DINOv2 CLS representation space using the Vendi Score \citep{friedman2023vendi} of the cosine-similarity matrix of L2-normalized CLS embeddings, and the mean pairwise Euclidean distance between CLS embeddings. Rewrite diversity is measured analogously in CLIP text-embedding space using Prompt Vendi and the mean pairwise cosine distance among rewritten prompts. Finally, CLIP-align, HPSv2 \citep{wu2023human}, and PickScore \citep{kirstain2023pick} assess the semantic compatibility and human-preference-style quality of each generated image relative to the input prompt. CLIP-align is the nonnegative cosine similarity between the L2-normalized CLIP embeddings of a generated image and its original prompt. Diversity metrics are computed within each group of rewrites or images generated for the same original prompt and then averaged across the evaluation set, following within-prompt conditional evaluation. For all metrics, higher values indicate greater diversity or stronger fidelity and preference, respectively.

\subsection{Experiment Details}
We use a frozen Stable Diffusion v1.4 generator \citep{rombach2022high} and optimize only the prompt-rewriting policy. For all baselines and reward functions, both Qwen3-0.6B(base) \citep{yang2025qwen3} and GPT-2 (124M) \citep{radford2019language} are already supervised fine-tuned on a shared set of curated prompt rewrites. We then train the models on 1,024 randomly sampled MS-COCO prompts \citep{lin2014coco} and evaluate them on disjoint MS-COCO and Lexica prompts \citep{shen2023prompt}. DARS uses $h=20$ for Gaussian kernels, and the alignment reward uses a CLIP-similarity threshold of $0.2$.

We compare DARS-GRPO with vanilla GRPO, PAG, and GFlowNet prompt optimization, together with Stable Diffusion v1.4 with random seed initialization and CADS \citep{sadat2024cads} as control baselines. All methods are evaluated using the same held-out prompts, generation budget, diffusion model, and sampling procedure. Diffusion seeds are sampled independently across methods, so image-diversity metrics also capture variation introduced by stochastic diffusion sampling while prompt-diversity metrics are unaffected by this confound.

\subsection{Results and Interpretation}

\begin{table*}[t]
\centering
\scriptsize
\caption{Prompt Adaptation for Text-to-Image Generation results on MS-COCO and Lexica \textit{(avg over 5 seeds)}.}
\label{tab:t2i_combined}
\resizebox{\textwidth}{!}{%
\begin{tabular}{llcccccccccccccc}
\toprule
&
&
\multicolumn{2}{c}{Image Vendi $\uparrow$} &
\multicolumn{2}{c}{CLIP-align $\uparrow$} &
\multicolumn{2}{c}{HPSv2 $\uparrow$} &
\multicolumn{2}{c}{PickScore $\uparrow$} &
\multicolumn{2}{c}{Prompt Vendi $\uparrow$} &
\multicolumn{2}{c}{Prompt diversity $\uparrow$} &
\multicolumn{2}{c}{DINO pairwise distance $\uparrow$}\\

\cmidrule(lr){3-4}
\cmidrule(lr){5-6}
\cmidrule(lr){7-8}
\cmidrule(lr){9-10}
\cmidrule(lr){11-12}
\cmidrule(lr){13-14}
\cmidrule(lr){15-16}

Policy LM & Method
& COCO & Lexica
& COCO & Lexica
& COCO & Lexica
& COCO & Lexica
& COCO & Lexica
& COCO & Lexica
& COCO & Lexica\\
\midrule

-- & SD v1.4
& 8.60 & 9.52
& 0.27 & 0.27
& 0.26 & 0.25
& 21.53 & 20.91
& -- & --
& -- & --
& 40.58 & 41.84\\

-- & CADS
& 12.93 & 14.08
& 0.26 & 0.26
& 0.24 & 0.23
& 20.86 & 20.28
& -- & --
& -- & --
& 46.22 & 46.96\\

\midrule
\multicolumn{16}{l}{\textit{Combined reward}}\\

Qwen3-0.6B & GRPO
& 8.76 & 9.81
& 0.24 & 0.24
& 0.23 & 0.23
& 20.92 & \textbf{20.69}
& 1.01 & 1.04
& 0.00 & 0.01
& 40.90 & 42.10\\

Qwen3-0.6B & DARS-GRPO
& \textbf{13.04} & \textbf{14.45}
& 0.24 & 0.24
& 0.23 & 0.23
& 20.80 & 20.37
& \textbf{1.43} & \textbf{1.67}
& \textbf{0.05} & 0.06
& \textbf{46.92} & \textbf{47.89}\\

Qwen3-0.6B & PAG
& 8.25 & 11.51
& 0.26 & \textbf{0.26}
& 0.25 & \textbf{0.24}
& 21.34 & \textbf{20.69}
& 1.26 & 1.59
& 0.04 & \textbf{0.07}
& 40.20 & 44.64\\

Qwen3-0.6B & GFN
& 8.74 & 11.69
& 0.26 & \textbf{0.26}
& 0.25 & \textbf{0.24}
& 21.31 & \textbf{20.69}
& 1.22 & 1.59
& 0.04 & \textbf{0.07}
& 40.50 & 44.71\\[1pt]

GPT-2 & GRPO
& 9.60 & 10.68
& 0.24 & \textbf{0.24}
& 0.23 & \textbf{0.23}
& 20.87 & \textbf{20.43}
& 1.09 & 1.18
& 0.01 & 0.03
& 41.85 & 42.87\\

GPT-2 & DARS-GRPO
& \textbf{12.72} & \textbf{14.67}
& 0.23 & 0.23
& 0.23 & \textbf{0.23}
& 20.68 & 20.26
& \textbf{1.63} & 1.78
& 0.08 & 0.10
& \textbf{47.75} & \textbf{48.60}\\

GPT-2 & PAG
& 9.22 & 14.03
& \textbf{0.26} & \textbf{0.24}
& \textbf{0.25} & \textbf{0.23}
& \textbf{21.16} & 20.35
& 1.61 & \textbf{2.14}
& \textbf{0.09} & \textbf{0.13}
& 41.36 & 46.43\\

GPT-2 & GFN
& 9.02 & 13.67
& 0.25 & \textbf{0.24}
& 0.24 & 0.22
& 21.12 & 20.26
& 1.15 & 1.69
& 0.02 & 0.08
& 41.03 & 46.67\\

\midrule
\multicolumn{16}{l}{\textit{Alignment-only reward}}\\

Qwen3-0.6B & GRPO
& 9.47 & 10.94
& 0.26 & \textbf{0.26}
& 0.25 & \textbf{0.24}
& 21.24 & 20.55
& 1.59 & 1.64
& 0.07 & 0.08
& 41.68 & 43.52\\

Qwen3-0.6B & DARS-GRPO
& \textbf{20.13} & \textbf{20.03}
& 0.22 & 0.21
& 0.22 & 0.22
& 20.34 & 19.81
& \textbf{2.43} & \textbf{2.40}
& \textbf{0.16} & \textbf{0.16}
& \textbf{56.23} & \textbf{55.76}\\

Qwen3-0.6B & PAG
& 9.53 & 13.88
& 0.25 & 0.24
& 0.25 & \textbf{0.24}
& 21.22 & 20.48
& 1.79 & 2.33
& 0.11 & 0.15
& 41.99 & 46.86\\

Qwen3-0.6B & GFN
& 8.93 & 12.13
& 0.26 & \textbf{0.26}
& 0.25 & \textbf{0.24}
& \textbf{21.27} & \textbf{20.60}
& 1.30 & 1.62
& 0.05 & 0.08
& 40.99 & 44.97\\[1pt]

GPT-2 & GRPO
& 8.76 & 10.22
& 0.26 & \textbf{0.26}
& \textbf{0.26} & \textbf{0.24}
& \textbf{21.42} & \textbf{20.66}
& 1.03 & 1.11
& 0.01 & 0.02
& 40.93 & 42.52\\

GPT-2 & DARS-GRPO
& \textbf{22.87} & \textbf{22.64}
& 0.23 & 0.22
& 0.23 & 0.22
& 20.58 & 19.89
& \textbf{2.15} & \textbf{2.29}
& \textbf{0.14} & \textbf{0.15}
& \textbf{55.57} & \textbf{55.00}\\

GPT-2 & PAG
& 8.51 & 13.15
& 0.25 & 0.24
& 0.25 & 0.23
& 21.23 & 20.38
& 1.20 & 1.66
& 0.03 & 0.07
& 40.55 & 45.78\\

GPT-2 & GFN
& 8.60 & 14.08
& 0.26 & 0.24
& 0.24 & 0.22
& 21.19 & 20.33
& 1.17 & 1.79
& 0.03 & 0.08
& 40.54 & 46.03\\

\bottomrule
\end{tabular}}
\end{table*}

Across both datasets, Policy LMs, and reward settings, DARS-GRPO consistently improves upon vanilla GRPO across all three diversity metrics (Table~\ref{tab:t2i_combined}) without significant loss of image fidelity. Further, among all prompt-adaptation methods, it achieves the highest DINO pairwise distances and Image Vendi scores, indicating that density-aware reward scaling broadens the learned rewrite distribution and increases coverage in DINOv2 image-representation space. The high Prompt Vendi and diversity scores further indicate that this image diversity arises from diverse prompt rewrites, while CLIP-align, HPSv2, and PickScore imply close image fidelity. 

Changing the reward function affects performance across all metrics, showing that DARS responds to the reward landscape rather than producing diversity gains randomly. Consistent with our hypothesis, the binary gated alignment-only reward yields broader sampling from uniformly scored reward zones, whereas the combined reward forces the policy to learn a more dynamic reward landscape characterized by aesthetically-driven peaks, thus, sampling from a more varied distribution as reflected by the higher average Image Vendi scores.

Having demonstrated the effectiveness of DARS on a real-world RL problem with an intractable reward-proportional distribution, we next study the effect of the kernel width used for local density estimation on the learned outcome distribution.


\FloatBarrier

\section{Ablation Study}

\begin{table}[t]
\centering
\small
\caption{Kernel-width ablation results \textit{(avg over 5 seeds)} }
\label{tab:ablations}
\resizebox{\columnwidth}{!}{%
\begin{tabular}{lcccc}
\toprule
Width & Image Vendi $\uparrow$ & CLIP-align $\uparrow$ & Prompt Diversity $\uparrow$ & Prompt Vendi $\uparrow$\\
\midrule
10 & 9.99 & 0.26 & 0.12 & 1.99\\
15 & 14.12 & 0.25 & 0.21 & 3.34\\
20 & 18.01 & 0.24 & 0.17 & 2.74\\
25 & 19.77 & 0.23 & 0.22 & 3.47\\
30 & 12.14 & 0.25 & 0.11 & 2.09\\
\bottomrule
\end{tabular}}
\end{table}

As discussed in \ref{behavior}, kernel width $h$ affects reward-scaling in DARS and hence needs to be calibrated for accurate density estimation. For the experiments in Section \ref{sec:t2i}, we demonstrate the direct effects of varying kernel widths on image-diversity produced by DARS (Table~\ref{tab:ablations}). 
Diversity metrics consistently peak at intermediate kernel widths and degrade when the bandwidth is either too small or too large, while image quality remains largely unchanged. This behavior reflects how DARS estimates local outcome density. 
With a very small kernel width, nearly every generated image appears isolated, causing density estimates to be dominated by self-similarity and resulting in almost uniform reward scaling. 
Conversely, an excessively large kernel width treats distant images as belonging to the same neighborhood, oversmoothing density estimates and diminishing the relative reward differences that encourage exploration. 
Intermediate kernel widths instead resolve meaningful local neighborhoods allowing for effective density estimation in DARS.

\section{Limitations}
DARS relies on a meaningful representation space and kernel bandwidth for estimating local outcome density. Its effectiveness may therefore degrade when embedding distances poorly reflect semantic similarity or when no single bandwidth captures the relevant neighborhood structure. The pairwise density computation also scales quadratically with rollout-group size, potentially limiting its use with large groups. Our experiments focus on a controlled navigation task and text-to-image prompt adaptation with fixed image representations and generators. Broader evaluation across RL domains, policy architectures, and learned representations is needed. Finally, our results establish empirical approximation to reward-proportional sampling rather than a general convergence guarantee.

\section{Conclusion}
We introduced Density-Aware Reward Scaling (DARS), a lightweight mechanism for promoting diverse, high-reward outcomes within standard policy-gradient training. By inversely scaling rewards using local outcome density, DARS counteracts repeated reinforcement of densely sampled regions without requiring a critic, auxiliary objective, learned density model, or modified policy-gradient loss. In a controlled multimodal navigation task, DARS-GRPO prevents the mode collapse observed with GRPO and closely recovers target mode allocation. In text-to-image prompt adaptation, it consistently increases prompt and image diversity across datasets, Policy LMs, and reward settings while retaining semantic fidelity. These results position density-aware reward scaling as a simple yet effective approach to improving outcome coverage.

\bibliography{references}



\newtheorem{theorem}{Theorem}
\newtheorem{proof}{Proof}
\newtheorem{corollary}{Corollary}

\setcounter{secnumdepth}{2}

\title{Density-Aware Reward Scaling for Diverse Prompt Adaptation in Text-to-Image Generation}

\author{Anonymous Submission}
\affiliations{}


\def\isDARSMainFile{1}

\appendix
\makeatletter
\@ifundefined{isDARSMainFile}{\newif\ifDARSStandalone\DARSStandalonetrue}{\newif\ifDARSStandalone\DARSStandalonefalse}
\makeatother

\ifDARSStandalone
\documentclass[10pt]{article}
\usepackage[margin=1in]{geometry}
\usepackage{amsmath,amssymb,amsthm,mathtools}
\usepackage{microtype}
\usepackage[hidelinks]{hyperref}
\newtheorem{theorem}{Theorem}
\title{Continuous-Space Justification of Density-Aware Reward Scaling}
\author{}
\date{}
\fi

\providecommand{\E}{\mathbb{E}}
\providecommand{\Var}{\operatorname{Var}}
\providecommand{\R}{\mathbb{R}}
\providecommand{\cR}{\mathcal{R}}
\providecommand{\op}{\mathrm{op}}

\ifDARSStandalone
\begin{document}
\maketitle
\fi

\section{Mathematical Foundations for DARS}
\subsection{Inverse policy-density scaling yields a reward-proportional policy}
\label{sec:inverse-density-foundation}

Let the continuous outcome space be $\mathcal{Y}\subseteq\R^d$. At training time $t$, let the policy induce a probability density $p_t(y)$ on $\mathcal{Y}$. Let the reward be a measurable function $r(y)\ge 0$ satisfying
\begin{equation}
0<\cR:=\int_{\mathcal{Y}} r(y)\,dy<\infty,
\end{equation}
and assume $p_t(y)>0$ wherever $r(y)>0$. The desired reward-proportional density is
\begin{equation}
p^\star(y)=\frac{r(y)}{\cR}.
\label{eq:target}
\end{equation}

We prove the continuous analogue of inverse probability scaling. Parameterize an unrestricted policy density by a logit function $f$,
\begin{equation}
p_f(y)=\frac{e^{f(y)}}{\int_{\mathcal{Y}}e^{f(u)}\,du}.
\end{equation}
At iteration $t$, freeze the current policy density in the reward denominator and use
\begin{equation}
\widetilde r_t(y)=\frac{r(y)}{p_t(y)}.
\label{eq:inverse-density-reward}
\end{equation}
For the population objective
\begin{equation}
J_t(f)=\int_{\mathcal{Y}}p_f(y)\widetilde r_t(y)\,dy,
\end{equation}
the following result holds.

\begin{theorem}
At the current policy $p_t=p_{f_t}$, the functional policy-gradient direction is
\begin{equation}
\boxed{
\frac{\delta J_t}{\delta f}(y)\bigg|_{f=f_t}
=r(y)-\cR p_t(y)
=\cR\bigl(p^\star(y)-p_t(y)\bigr).
}
\label{eq:functional-gradient}
\end{equation}
Hence the reward-proportional density $p^\star(y)=r(y)/\cR$ is the unique stationary probability density, interpreted in the closure of the exponential-family parameterization when $r$ vanishes on part of the outcome space. Under the corresponding population gradient flow, every convergent density limit is $p^\star$.
\end{theorem}

\begin{proof}
For any perturbation $v$, the normalized exponential density obeys
\begin{equation}
\frac{d}{ds}p_{f_t+sv}(y)\bigg|_{s=0}
=p_t(y)\left(v(y)-\E_{p_t}[v]\right).
\end{equation}
Because the denominator in Equation~\eqref{eq:inverse-density-reward} is stop-gradient,
\begin{align}
\frac{d}{ds}J_t(f_t+sv)\bigg|_{s=0}
&=\int_{\mathcal{Y}}
\frac{r(y)}{p_t(y)}p_t(y)\\
&\qquad\times\left(v(y)-\E_{p_t}[v]\right)\,dy\\
&=\int_{\mathcal{Y}}r(y)v(y)\,dy\\
&\qquad-\cR\int_{\mathcal{Y}}p_t(y)v(y)\,dy\\
&=\int_{\mathcal{Y}}\!\left(r(y)-\cR p_t(y)\right)v(y)\,dy.
\end{align}
Since this identity holds for every perturbation $v$, Equation~\eqref{eq:functional-gradient} follows. A stationary density must satisfy
\begin{equation}
r(y)-\cR p_t(y)=0
\end{equation}
almost everywhere, and therefore
\begin{equation}
p_t(y)=\frac{r(y)}{\cR}=p^\star(y).
\end{equation}

The direction is not merely stationary at $p^\star$; it points toward it. Consider the population gradient flow
\begin{equation}
\partial_t f_t(y)=\cR\bigl(p^\star(y)-p_t(y)\bigr)
\end{equation}
and the functional
\begin{equation}
\Phi(f)=\log\!\int_{\mathcal{Y}}e^{f(y)}dy
-\int_{\mathcal{Y}}p^\star(y)f(y)dy.
\end{equation}
Its functional derivative is $\delta\Phi/\delta f=p_f-p^\star$. Therefore, along the gradient flow,
\begin{align}
\frac{d}{dt}\Phi(f_t)
&=\int_{\mathcal{Y}}
\bigl(p_t(y)-p^\star(y)\bigr)\partial_t f_t(y)\,dy\\
&=-\cR\int_{\mathcal{Y}}
\bigl(p_t(y)-p^\star(y)\bigr)^2dy\le 0.
\end{align}
Equality holds only when $p_t=p^\star$ almost everywhere. Thus the only stationary density, and hence any convergent population-gradient-flow density limit, is the reward-proportional density in Equation~\eqref{eq:target}.
\end{proof}

The cancellation underlying the result can also be seen directly. A small region $dy$ around $y$ is sampled with probability approximately $p_t(y)dy$. Its per-sample scaled reward is $r(y)/p_t(y)$. Thus its aggregate expected learning weight is proportional to
\begin{equation}
\underbrace{p_t(y)dy}_{\text{sampling probability}}
\underbrace{\frac{r(y)}{p_t(y)}}_{\text{inverse-density reward}}
=r(y)dy.
\end{equation}
The current policy density cancels, leaving learning mass proportional only to reward mass.

\subsection{The Gaussian kernel density score is proportional to the policy density}

Draw a rollout group of $G$ outcomes independently from the current policy,
\begin{equation}
Y_1,\ldots,Y_G\overset{\mathrm{iid}}{\sim}p_t.
\end{equation}
DARS uses the unnormalized Gaussian kernel
\begin{equation}
K_h(y,z)=\exp\!\left(-\frac{\|y-z\|^2}{2h^2}\right).
\end{equation}
Instead of separating any sample from the rest of the group, define the density-score field produced by the entire rollout group,
\begin{equation}
\rho_G(y)=\sum_{j=1}^{G}K_h(y,Y_j).
\label{eq:rho-field}
\end{equation}
The score assigned to a sampled outcome is simply
\begin{equation}
\rho_i=\rho_G(Y_i).
\label{eq:rho-i}
\end{equation}
Thus the same group of all $G$ policy samples is used throughout.

Define
\begin{equation}
C_h=(2\pi)^{d/2}h^d,
\qquad
\phi_h(z)=\frac{1}{C_h}
\exp\!\left(-\frac{\|z\|^2}{2h^2}\right).
\end{equation}
Then $\phi_h$ is the $\mathcal{N}(0,h^2I_d)$ density and
\begin{equation}
\widehat p_{G,h}(y)
:=
\frac{\rho_G(y)}{GC_h}
=
\frac{1}{G}\sum_{j=1}^{G}\phi_h(y-Y_j).
\label{eq:kde}
\end{equation}
Equation~\eqref{eq:kde} is exactly the Gaussian kernel-density estimator formed from the full rollout group.

\begin{theorem}[Gaussian density approximation]
For every fixed $y$,
\begin{equation}
\boxed{
\E[\widehat p_{G,h}(y)]
=
(p_t*\phi_h)(y),
}
\label{eq:mean-kde}
\end{equation}
where
\begin{equation}
(p_t*\phi_h)(y)
=
\int_{\mathcal{Y}}p_t(u)\phi_h(y-u)\,du.
\end{equation}
Assume that, on the region of interest, $p_t$ is twice continuously differentiable,
\begin{equation}
\sup_y\|\nabla^2p_t(y)\|_{\op}\le M,
\end{equation}
and $p_t(y)\le P$. If the region has a boundary, assume either that $p_t$ admits a $C^2$ extension to a neighborhood of that region or that the statement is restricted to interior points at the bandwidth scale. Then
\begin{equation}
\boxed{
\left|\E[\widehat p_{G,h}(y)]-p_t(y)\right|
\le \frac{dM}{2}h^2,
}
\label{eq:bias}
\end{equation}
and
\begin{equation}
\boxed{
\Var(\widehat p_{G,h}(y))
\le
\frac{P}{2^d\pi^{d/2}Gh^d}.
}
\label{eq:variance}
\end{equation}
Consequently, for a fixed query point $y$ and any $\delta\in(0,1)$, with probability at least $1-\delta$,
\begin{equation}
\boxed{
\left|\widehat p_{G,h}(y)-p_t(y)\right|
\le
\frac{dM}{2}h^2
+
\sqrt{
\frac{P}{
2^d\pi^{d/2}Gh^d\delta
}
}.
}
\label{eq:pointwise-bound}
\end{equation}
\end{theorem}

\begin{proof}
Using the full group in Equation~\eqref{eq:kde} and linearity of expectation,
\begin{align}
\E[\widehat p_{G,h}(y)]
&=
\E\left[
\frac{1}{G}\sum_{j=1}^{G}\phi_h(y-Y_j)
\right]\\
&=
\frac{1}{G}\sum_{j=1}^{G}
\E[\phi_h(y-Y_j)]\\
&=
\int_{\mathcal{Y}}
\phi_h(y-u)p_t(u)\,du\\
&=
(p_t*\phi_h)(y).
\end{align}
No element of the rollout group is separated from the others.

Let $Z\sim\mathcal{N}(0,h^2I_d)$. Then, away from boundary effects as specified above,
\begin{equation}
(p_t*\phi_h)(y)=\E[p_t(y-Z)].
\end{equation}
A second-order Taylor expansion gives
\begin{equation}
p_t(y-Z)
=
p_t(y)
-
\nabla p_t(y)^\top Z
+
\frac{1}{2}
Z^\top\nabla^2p_t(\xi)Z
\end{equation}
for a point $\xi$ on the segment between $y$ and $y-Z$. Since $\E[Z]=0$ and
\begin{equation}
\E\|Z\|^2=dh^2,
\end{equation}
we obtain
\begin{equation}
\left|(p_t*\phi_h)(y)-p_t(y)\right|
\le
\frac{M}{2}\E\|Z\|^2
=
\frac{dM}{2}h^2,
\end{equation}
which proves Equation~\eqref{eq:bias}.

For fixed $y$, the $G$ terms in Equation~\eqref{eq:kde} are independent and identically distributed. Hence
\begin{align}
\Var(\widehat p_{G,h}(y))
&=
\frac{1}{G}
\Var(\phi_h(y-Y))\\
&\le
\frac{1}{G}
\E[\phi_h(y-Y)^2]\\
&\le
\frac{P}{G}
\int_{\R^d}\phi_h(z)^2\,dz.
\end{align}
For the Gaussian kernel,
\begin{equation}
\int_{\R^d}\phi_h(z)^2\,dz
=
\frac{1}{2^d\pi^{d/2}h^d},
\end{equation}
which proves Equation~\eqref{eq:variance}. Chebyshev's inequality together with Equation~\eqref{eq:bias} gives Equation~\eqref{eq:pointwise-bound}.
\end{proof}

The preceding theorem shows directly how the approximation depends on bandwidth and group size:
\begin{equation}
\boxed{
\widehat p_{G,h}(y)
=
p_t(y)
+
O(h^2)
+
O_p\!\left((Gh^d)^{-1/2}\right).
}
\label{eq:kde-rate}
\end{equation}
The $O(h^2)$ term is the Gaussian smoothing bias, while the stochastic error decreases as the effective number of samples in an $h$-scale neighborhood, $Gh^d$, increases.

Equation~\eqref{eq:pointwise-bound} is stated for a fixed query point because that is the elementary finite-sample calculation. To evaluate the same full-group estimator at the sampled outcomes $Y_i$ without separating any sample, use the standard uniform-consistency property of Gaussian kernel-density estimators. On a compact region $\mathcal{D}$ satisfying the regularity assumptions above,
\begin{equation}
\sup_{y\in\mathcal{D}}
\left|
\widehat p_{G,h}(y)-p_t(y)
\right|
=
O(h^2)
+
O_p\!\left(
\sqrt{\frac{\log G}{Gh^d}}
\right),
\label{eq:uniform-rate}
\end{equation}
provided $h\to0$ and $Gh^d/\log G\to\infty$. Because Equation~\eqref{eq:uniform-rate} holds simultaneously for every $y\in\mathcal{D}$, it also holds when $y$ is any of the observed outcomes $Y_i\in\mathcal{D}$. Therefore,
\begin{equation}
\boxed{
\frac{\rho_i}{GC_h}
=
p_t(Y_i)
+
O(h^2)
+
O_p\!\left(
\sqrt{\frac{\log G}{Gh^d}}
\right).
}
\label{eq:sample-point-rate}
\end{equation}

In particular, on any region where $p_t(y)\ge m>0$, if $h$ and $G$ are chosen so that the right-hand error in Equation~\eqref{eq:sample-point-rate} is small relative to $m$, then
\begin{equation}
\boxed{
\rho_i
=
GC_h\,p_t(Y_i)(1+\varepsilon_i),
\qquad
|\varepsilon_i|\ll1.
}
\label{eq:proportional}
\end{equation}
Thus the DARS Gaussian score is approximately the policy probability density multiplied by the same global factor $GC_h$ for all outcomes in the region of interest.

The bandwidth tradeoff is now explicit. The bandwidth should be small enough that
\begin{equation}
p_t*\phi_h\approx p_t,
\end{equation}
but large enough that
\begin{equation}
Gh^d
\end{equation}
is not too small. In the standard regime
\begin{equation}
h\to0,
\qquad
\frac{Gh^d}{\log G}\to\infty,
\end{equation}
the approximation error vanishes uniformly on the region of interest.

Finally, Equation~\eqref{eq:proportional} implies
\begin{equation}
\frac{R_i}{\rho_i}
=
\frac{1}{GC_h}
\frac{R_i}{p_t(Y_i)}
\frac{1}{1+\varepsilon_i}.
\end{equation}
Thus, when the Gaussian kernel-density estimate is accurate, DARS differs from exact inverse policy-density scaling only by a common positive factor and a small multiplicative estimation error. Section~\ref{sec:inverse-density-foundation} then gives the corresponding reward-proportional population target.

\ifDARSStandalone
\end{document}
\fi

\FloatBarrier
\section{Navigation Task}

\begin{figure}[t]
    \centering
    \includegraphics[width=0.5\textwidth]{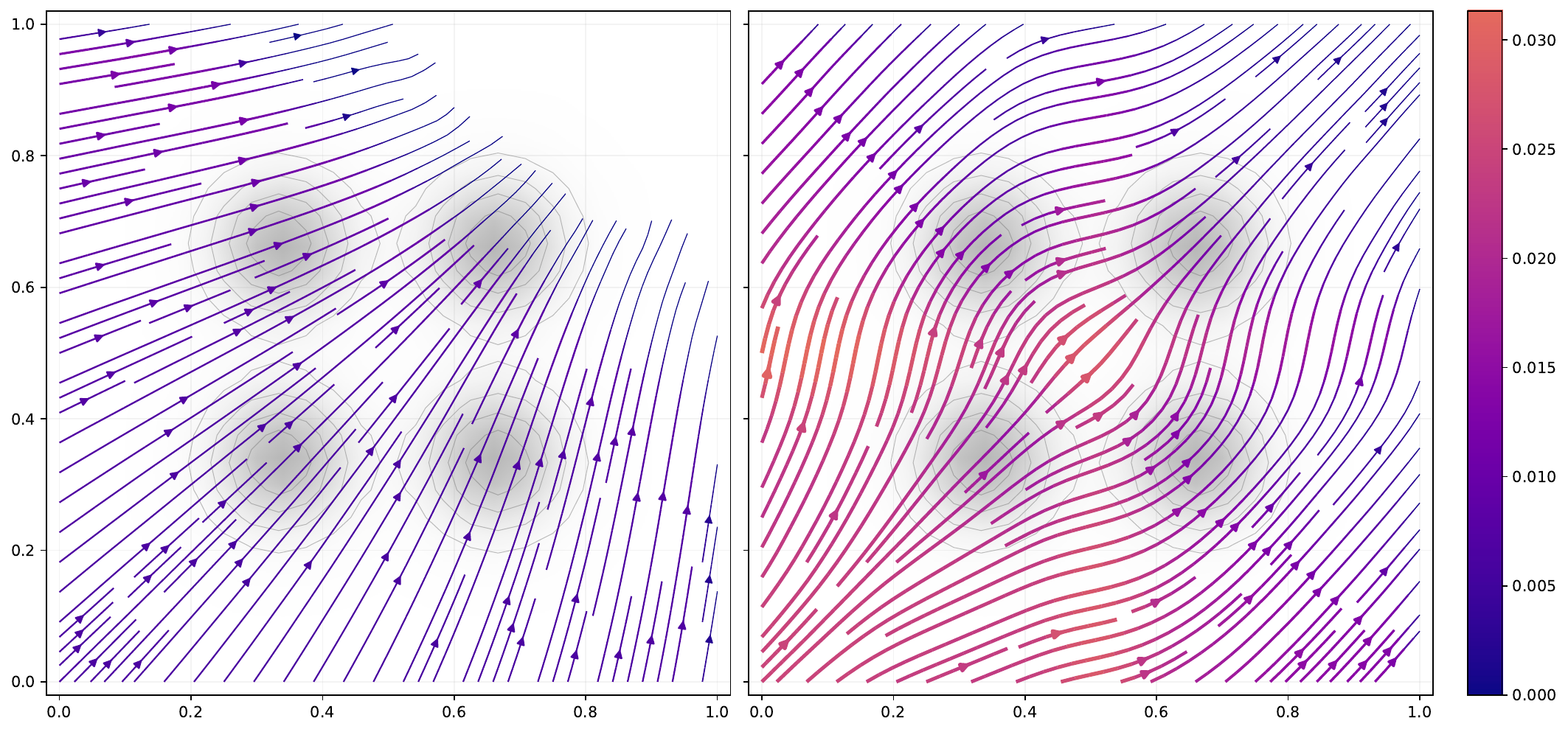}
    \caption{Policy-induced movement fields for GRPO (left) and DARS-GRPO (right) on the four-mode navigation task. Streamlines show the estimated mean one-step displacement, color indicates its magnitude, and gray contours show the normalized reward landscape.}
    \label{fig:navigation-force-field}
\end{figure}

Figure~\ref{fig:navigation-force-field} visualizes the movement fields induced
by the trained GRPO and DARS-GRPO policies. At each state
$s=(x,y)\in[0,1]^2$, actions are sampled repeatedly from the policy and used
to estimate the mean one-step displacement
$\mathbb{E}[s_{t+1}-s_t\mid s_t=s]$. The streamlines indicate its direction,
their color indicates its magnitude, and the gray contours show the normalized
reward landscape.

Terminal actions are treated as zero displacement. Consequently, a small
field magnitude may result from small proposed movements, frequent
termination, or cancellation among sampled actions.
The DARS-GRPO opptimised the policy such that the coverage over all the modes could be optimised in proportion to the reward.
The visualization thus
shows how density-aware reward scaling changes the learned navigation dynamics.
It should not be interpreted as a gradient or parameter-update field.

\section{Text-to-Image Prompt Adaptation}

\begin{figure*}[t]
    \centering
    \begin{minipage}[t]{0.86\textwidth}
        \centering
        \textbf{GPT-2}\par\smallskip
        \includegraphics[width=\linewidth]{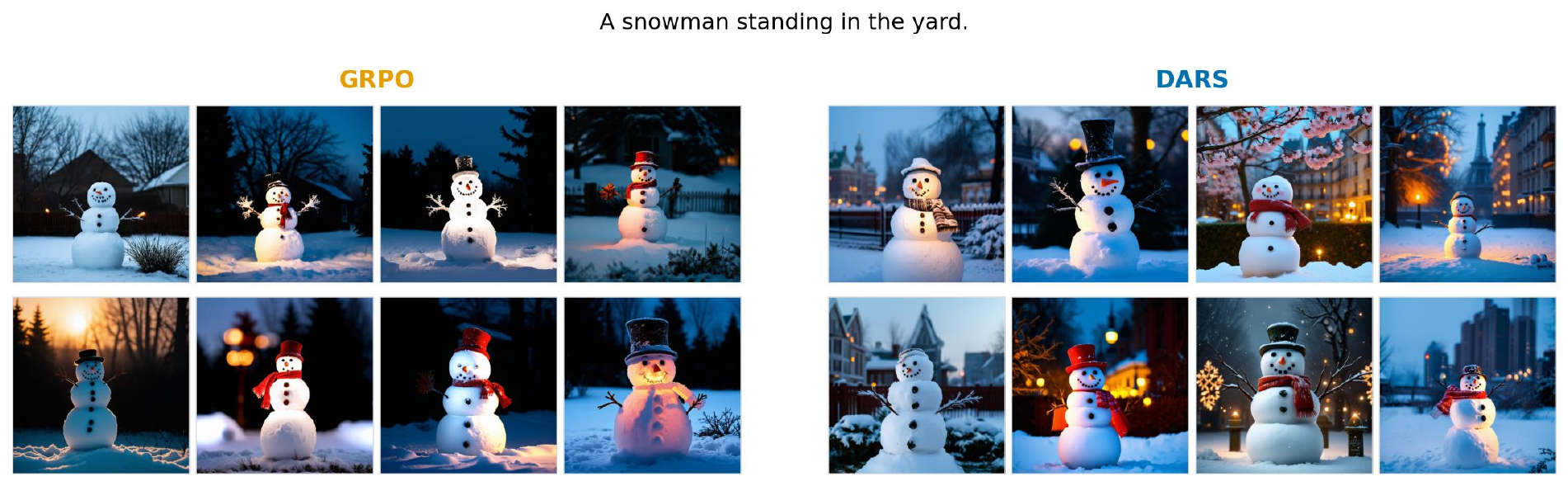}
    \end{minipage}\par\vspace{0.4em}
    \begin{minipage}[t]{0.86\textwidth}
        \centering
        \textbf{Qwen3-0.6B}\par\smallskip
        \includegraphics[width=\linewidth]{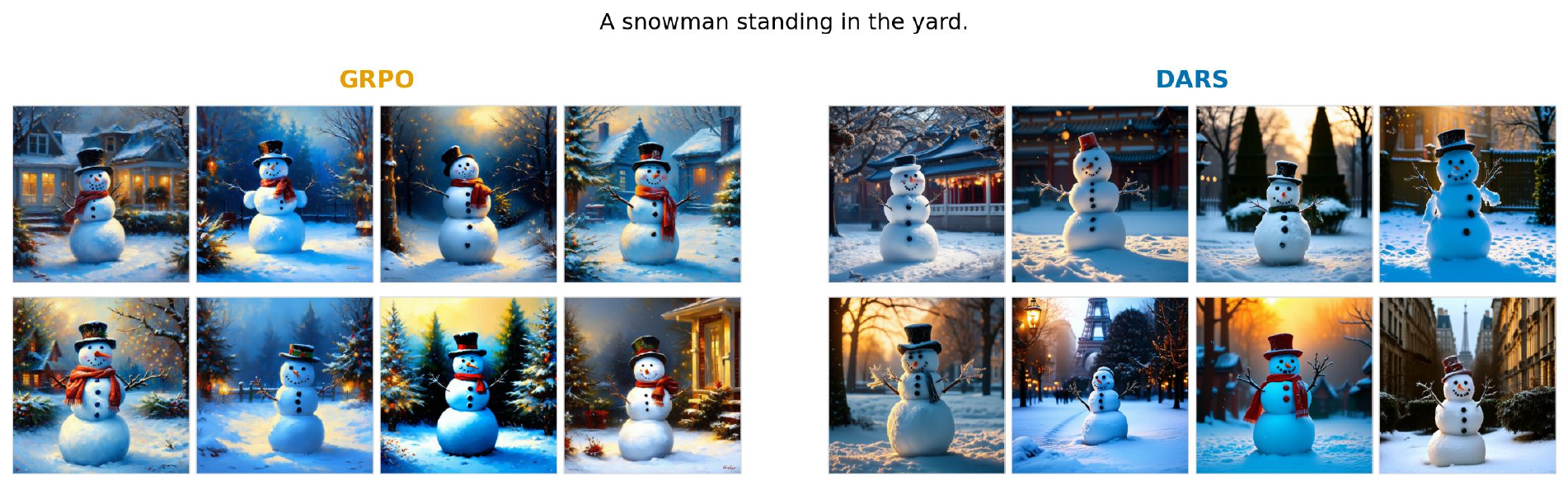}
    \end{minipage}

    \vspace{0.5em}

    \begin{minipage}[t]{0.86\textwidth}
        \centering
        \textbf{GPT-2}\par\smallskip
        \includegraphics[width=\linewidth]{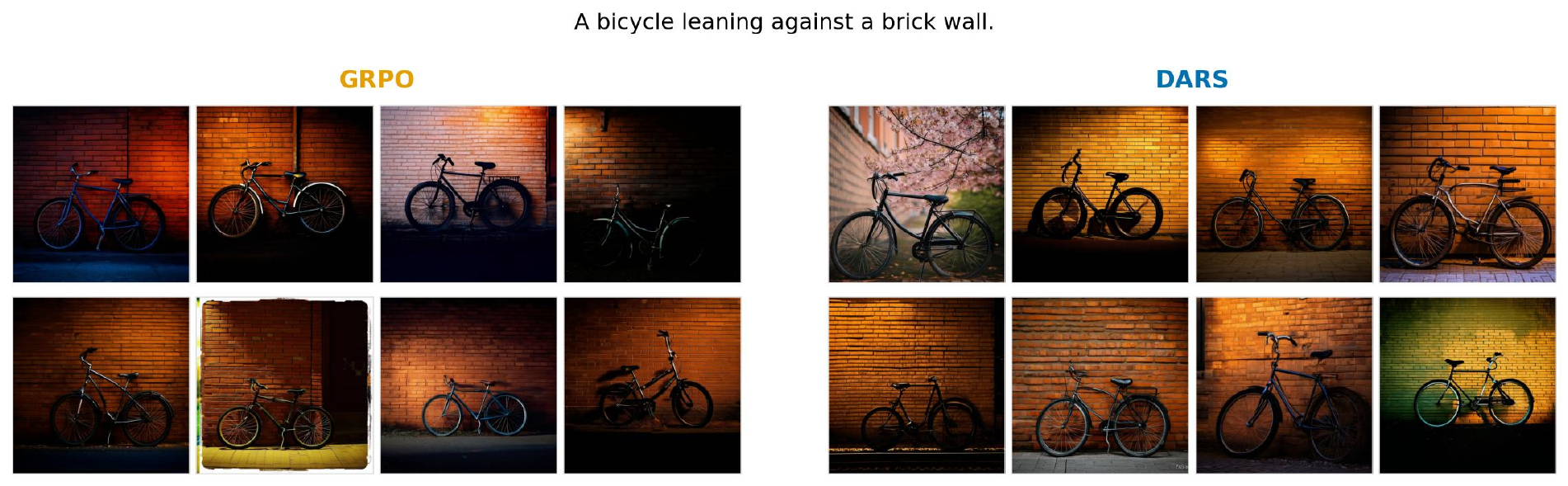}
    \end{minipage}\par\vspace{0.4em}
    \begin{minipage}[t]{0.86\textwidth}
        \centering
        \textbf{Qwen3-0.6B}\par\smallskip
        \includegraphics[width=\linewidth]{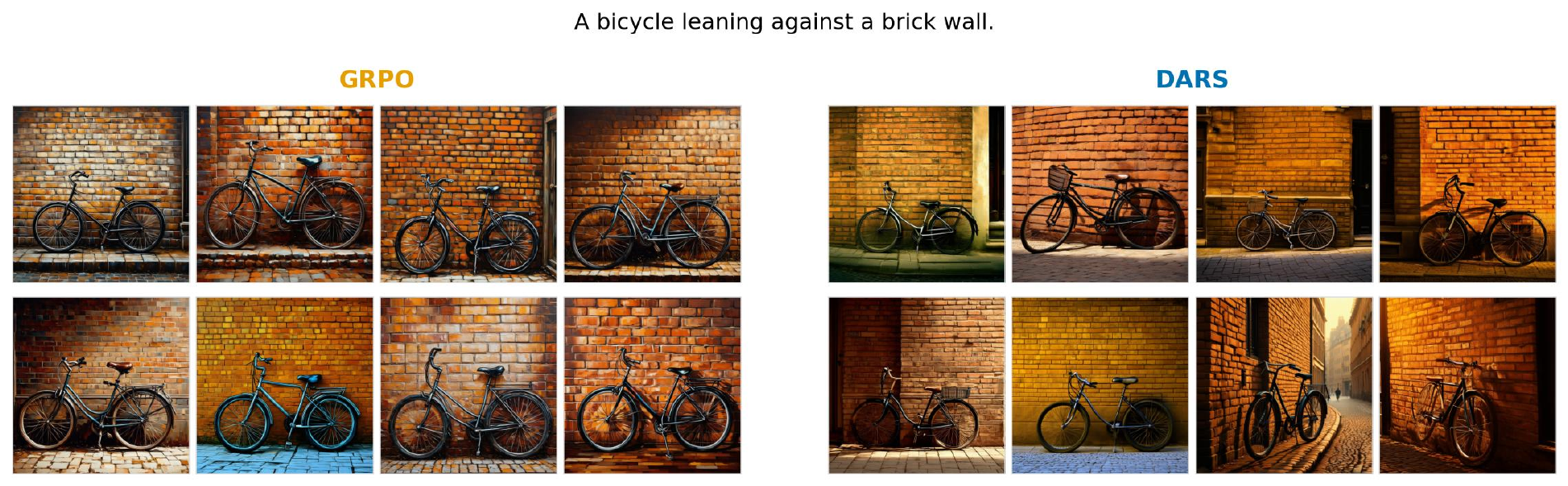}
    \end{minipage}
    \caption{Policy LM comparison for ``A snowman standing in the yard'' (top) and ``A bicycle leaning against a brick wall'' (bottom). Each embedded grid compares GRPO with DARS for the indicated language-model.}
    \label{fig:lm-comparison-01}
\end{figure*}

\begin{figure*}[t]
    \centering
    \begin{minipage}[t]{0.86\textwidth}
        \centering
        \textbf{GPT-2}\par\smallskip
        \includegraphics[width=\linewidth]{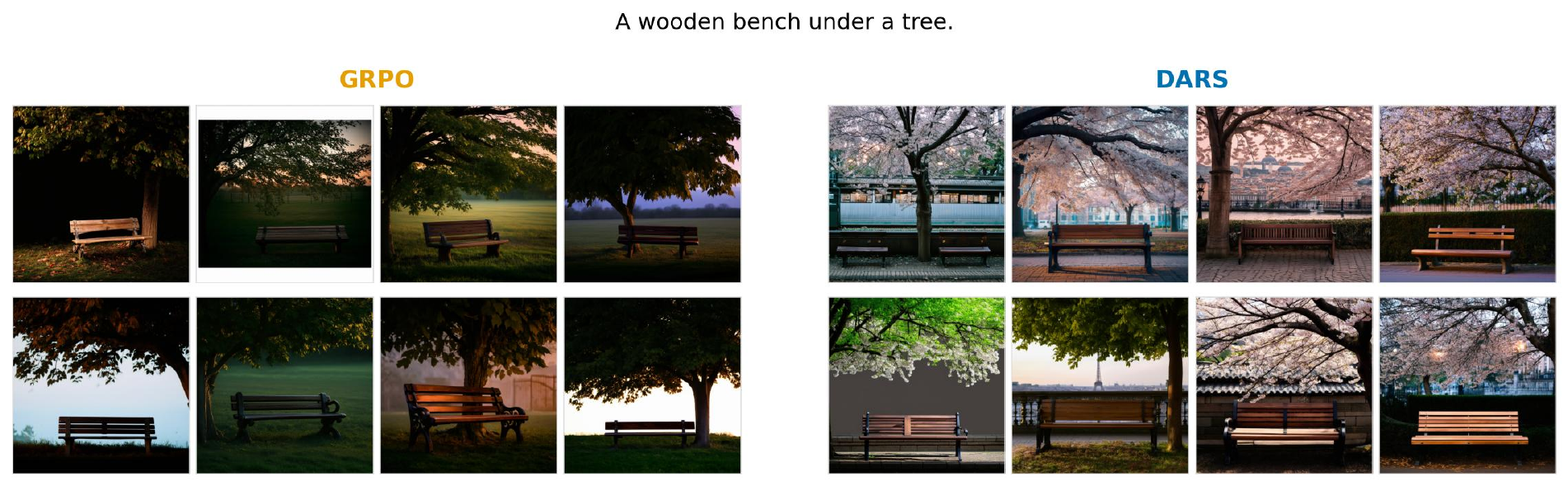}
    \end{minipage}\par\vspace{0.4em}
    \begin{minipage}[t]{0.86\textwidth}
        \centering
        \textbf{Qwen3-0.6B}\par\smallskip
        \includegraphics[width=\linewidth]{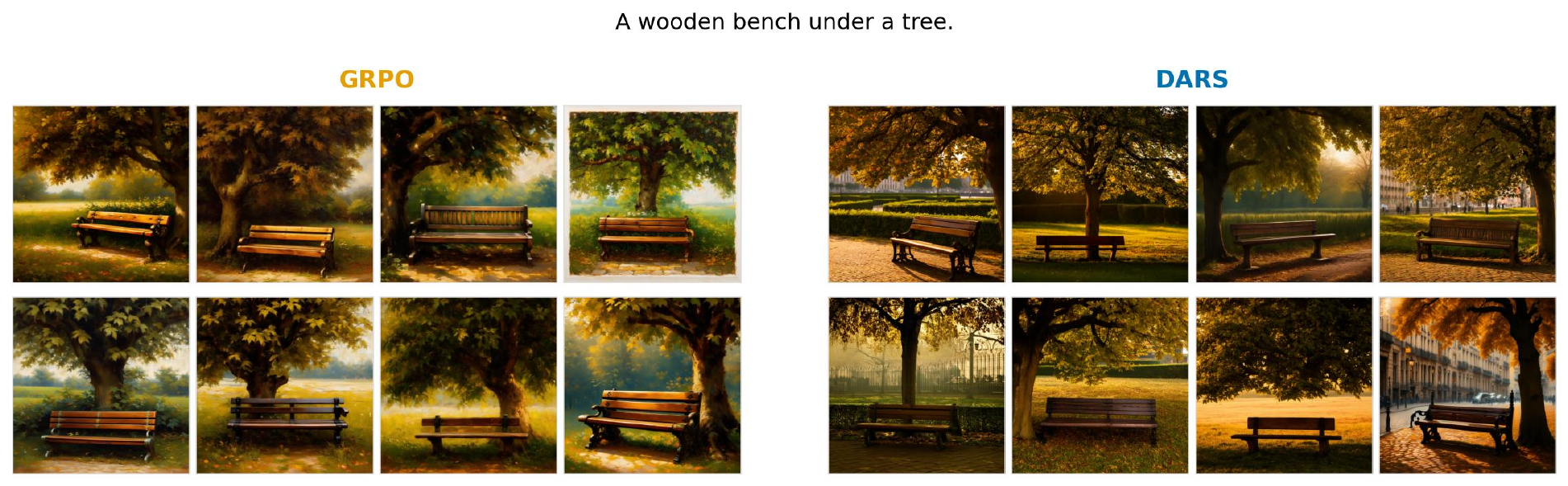}
    \end{minipage}

    \vspace{0.5em}

    \begin{minipage}[t]{0.86\textwidth}
        \centering
        \textbf{GPT-2}\par\smallskip
        \includegraphics[width=\linewidth]{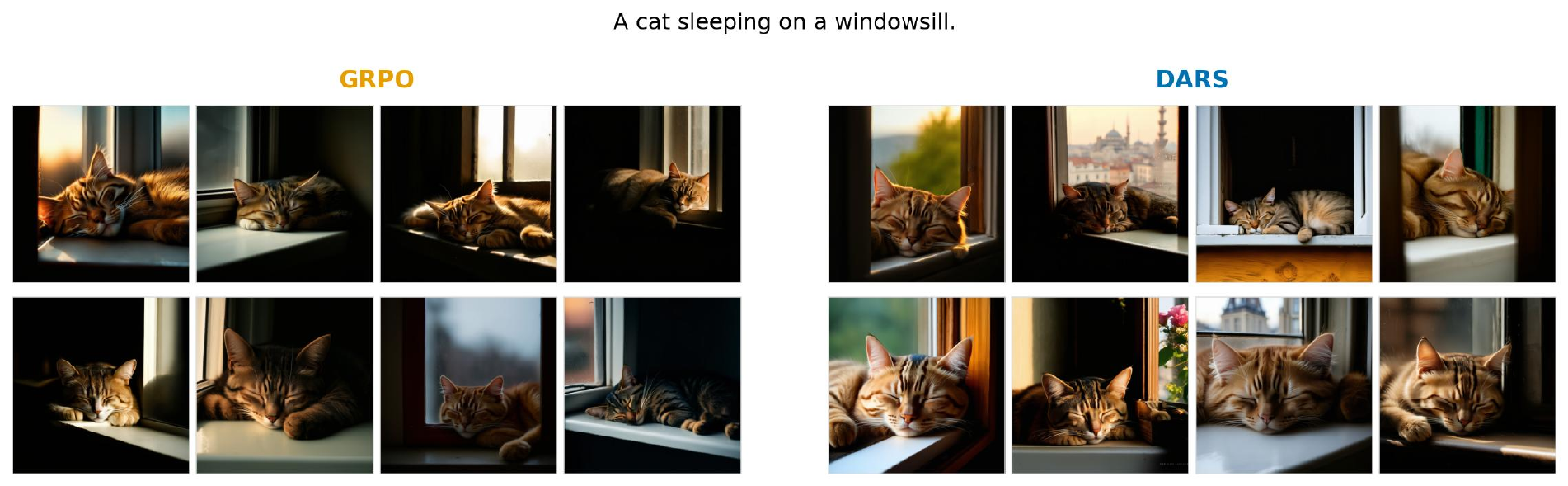}
    \end{minipage}\par\vspace{0.4em}
    \begin{minipage}[t]{0.86\textwidth}
        \centering
        \textbf{Qwen3-0.6B}\par\smallskip
        \includegraphics[width=\linewidth]{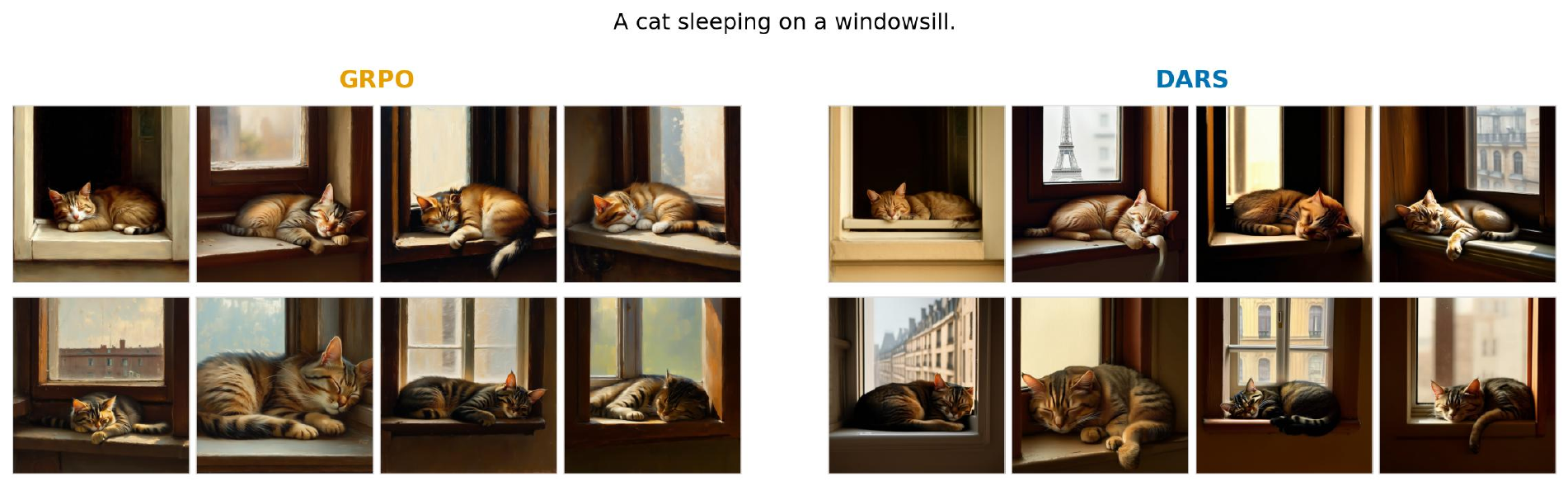}
    \end{minipage}
    \caption{Policy LM comparison for ``A wooden bench under a tree'' (top) and ``A cat sleeping on a windowsill'' (bottom). DARS expands the visual contexts while preserving the object and scene requested by each prompt.}
    \label{fig:lm-comparison-23}
\end{figure*}

\begin{figure*}[t]
    \centering
    \begin{minipage}[t]{0.86\textwidth}
        \centering
        \textbf{GPT-2}\par\smallskip
        \includegraphics[width=\linewidth]{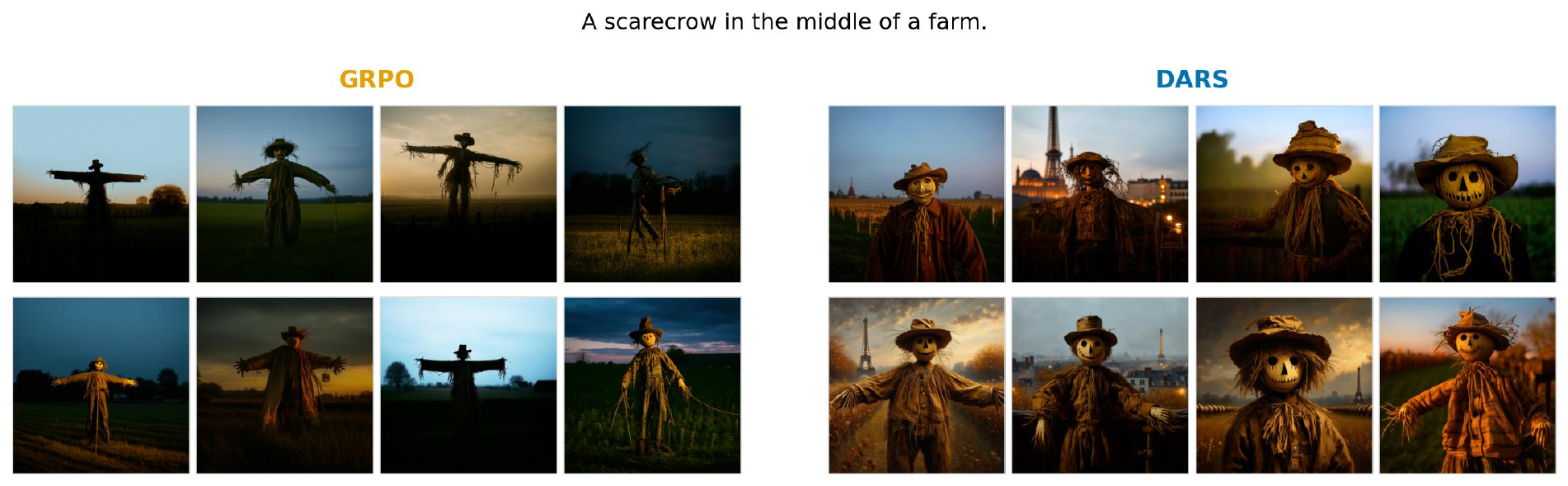}
    \end{minipage}\par\vspace{0.4em}
    \begin{minipage}[t]{0.86\textwidth}
        \centering
        \textbf{Qwen3-0.6B}\par\smallskip
        \includegraphics[width=\linewidth]{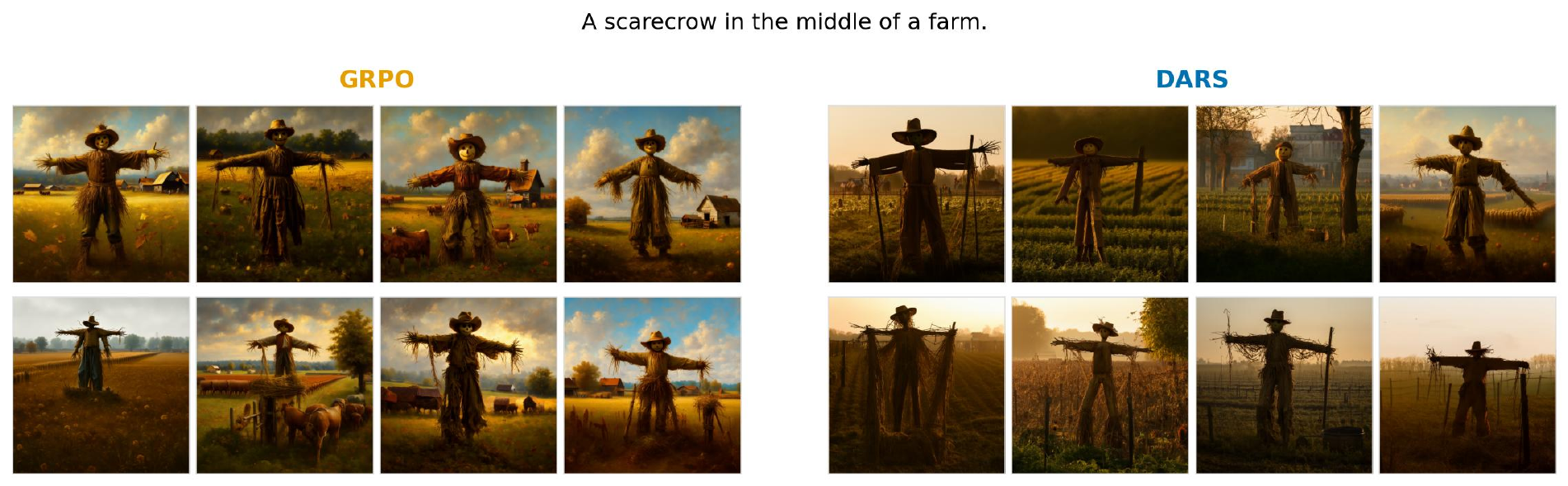}
    \end{minipage}

    \vspace{0.5em}

    \begin{minipage}[t]{0.86\textwidth}
        \centering
        \textbf{GPT-2}\par\smallskip
        \includegraphics[width=\linewidth]{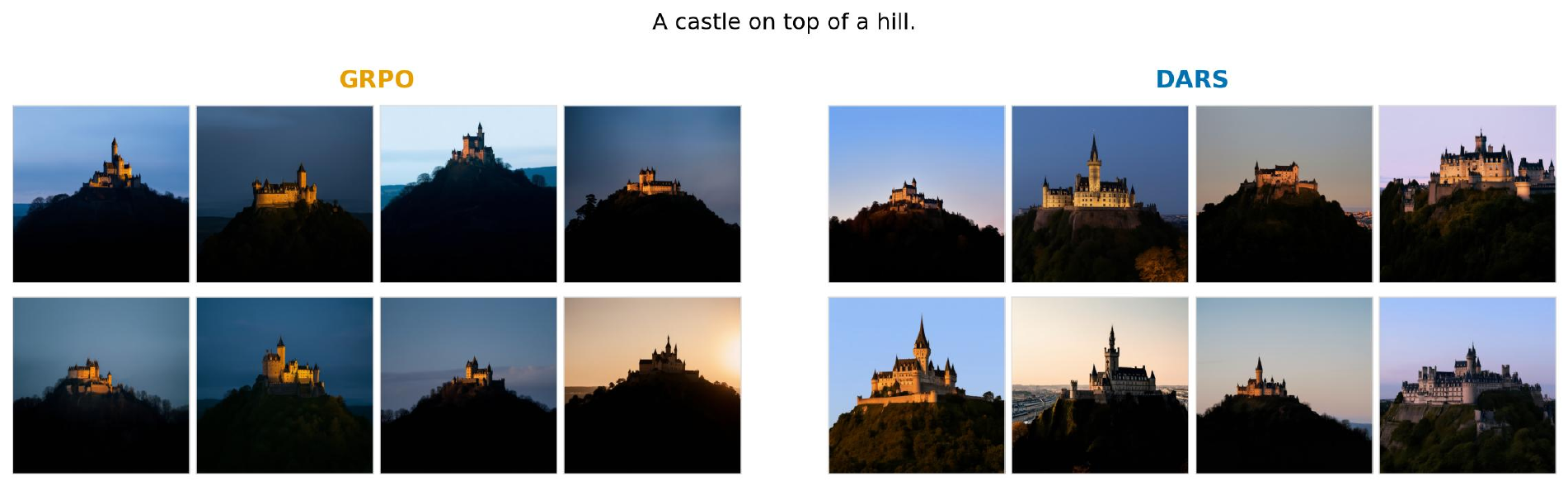}
    \end{minipage}\par\vspace{0.4em}
    \begin{minipage}[t]{0.86\textwidth}
        \centering
        \textbf{Qwen3-0.6B}\par\smallskip
        \includegraphics[width=\linewidth]{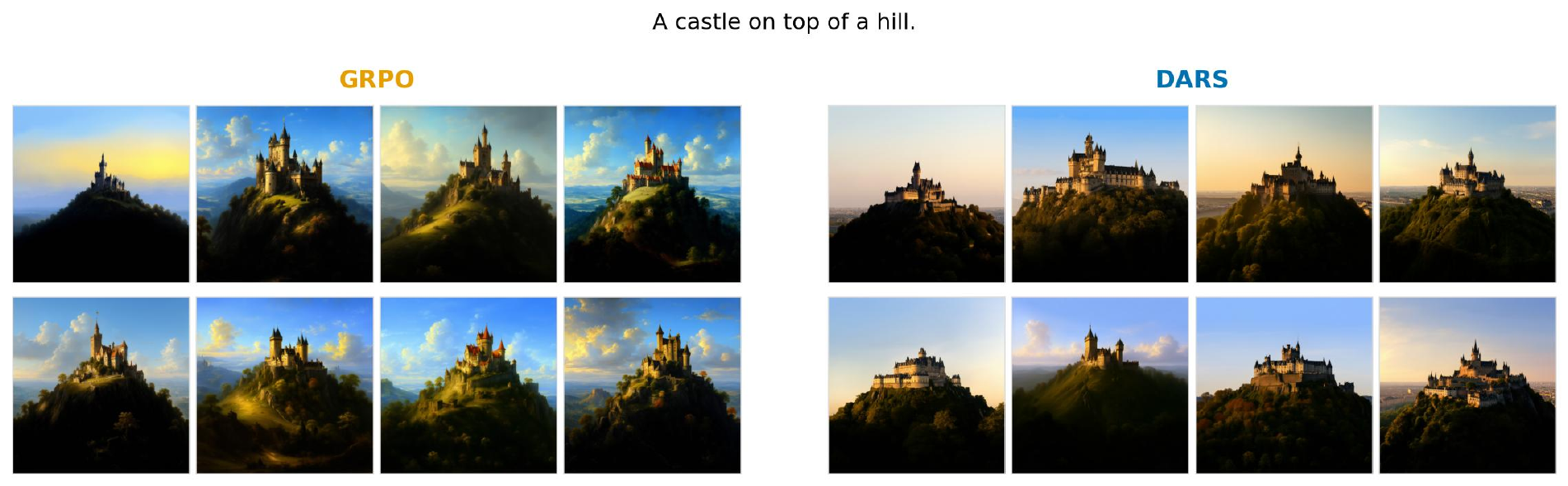}
    \end{minipage}
    \caption{Policy LM comparison for ``A scarecrow in the middle of a farm'' (top) and ``A castle on top of a hill'' (bottom). The DARS samples show greater variation in character appearance, architecture, framing, and illumination than the corresponding GRPO samples.}
    \label{fig:lm-comparison-45}
\end{figure*}

\begin{figure*}[t]
    \centering
    \begin{minipage}[t]{0.86\textwidth}
        \centering
        \textbf{GPT-2}\par\smallskip
        \includegraphics[width=\linewidth]{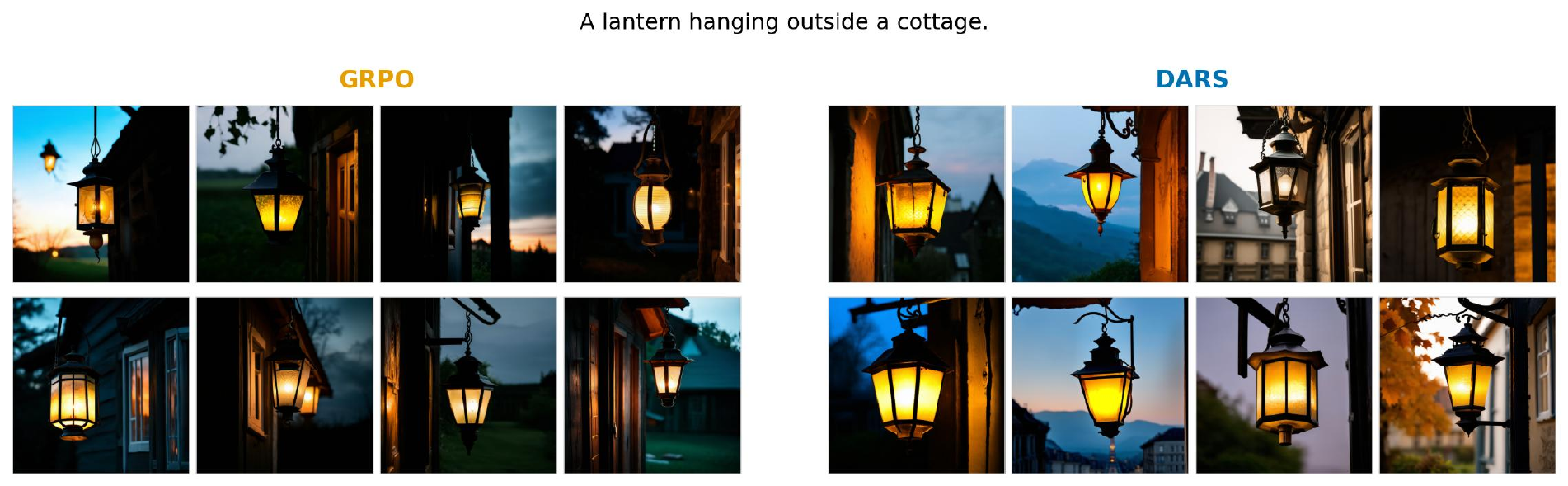}
    \end{minipage}\par\vspace{0.4em}
    \begin{minipage}[t]{0.86\textwidth}
        \centering
        \textbf{Qwen3-0.6B}\par\smallskip
        \includegraphics[width=\linewidth]{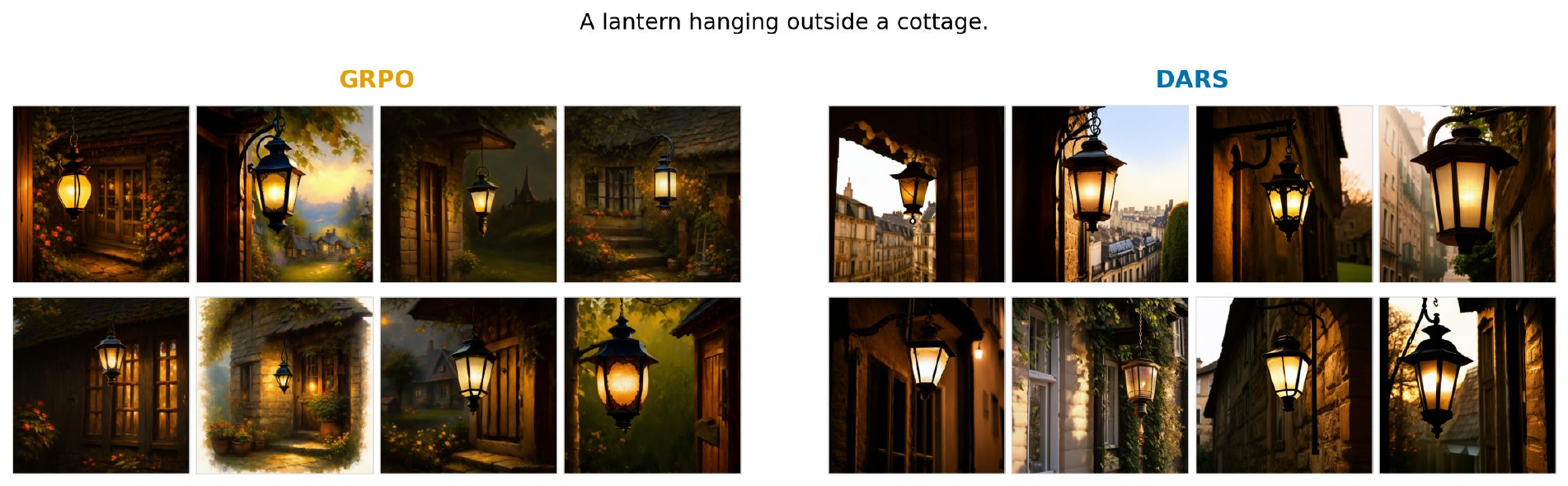}
    \end{minipage}

    \vspace{0.5em}

    \begin{minipage}[t]{0.86\textwidth}
        \centering
        \textbf{GPT-2}\par\smallskip
        \includegraphics[width=\linewidth]{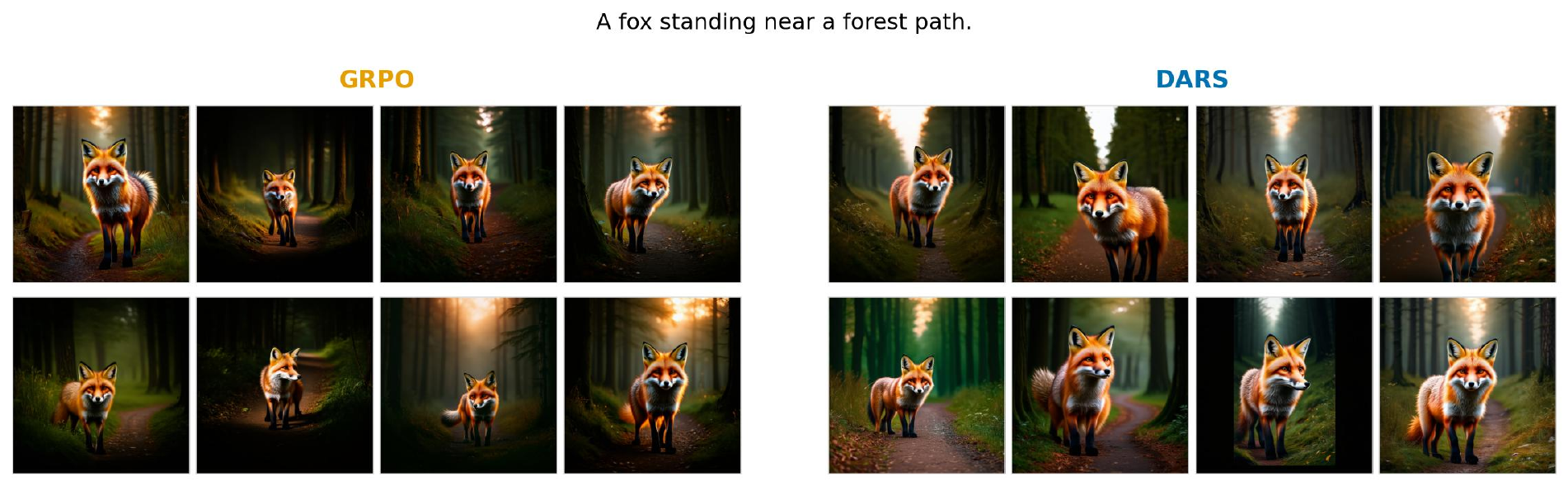}
    \end{minipage}\par\vspace{0.4em}
    \begin{minipage}[t]{0.86\textwidth}
        \centering
        \textbf{Qwen3-0.6B}\par\smallskip
        \includegraphics[width=\linewidth]{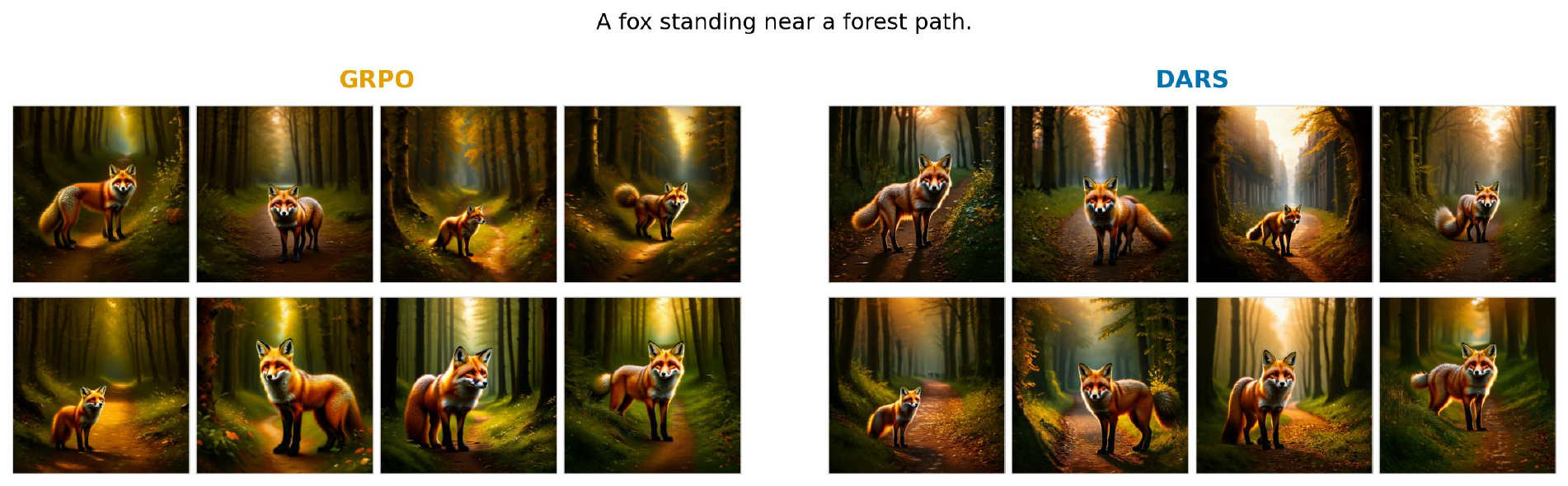}
    \end{minipage}
    \caption{Policy LM comparison for ``A lantern hanging outside a cottage'' (top) and ``A fox standing near a forest path'' (bottom). Across both policy LMs, DARS broadens the within-prompt visual distribution while maintaining the prompt's central subject.}
    \label{fig:lm-comparison-67}
\end{figure*}

\textbf{LM model choice: GPT-2 vs Qwen3-0.6B}
Figures~\ref{fig:lm-comparison-01}--\ref{fig:lm-comparison-67} compare GPT-2 and Qwen3-0.6B as policy LMs under the combined aesthetic and alignment reward. For each original prompt, GRPO generations are shown on the left and DARS generations on the right while keeping the diffusion model, sampling budget, and evaluation protocol fixed. This isolates the effect of the language-model used from that of the reinforcement learning objective.

Across both policy LMs, DARS consistently expands the diversity of generated images while preserving the semantic content of the original prompt. Compared with GRPO, DARS produces greater variation in scene layout, viewpoint, lighting, object appearance, and background context without substantially altering the primary subject. Although GPT-2 tends to make larger structural changes and Qwen3-0.6B generally preserves a more stable object identity, both exhibit the same qualitative benefit from density-aware reward scaling. These examples illustrate that the improvements obtained by DARS are not tied to a particular language model and generalize across different policy LMs.

\textbf{Reward choice: Combined vs Alignment-only}

\begin{figure*}[t]
    \centering
    \begin{minipage}[t]{0.86\textwidth}
        \centering
        \textbf{Combined}\par\smallskip
        \includegraphics[width=\linewidth]{Img_results/qwen_aes/prompt_000_rewrite_grid.pdf}
    \end{minipage}\par\vspace{0.4em}
    \begin{minipage}[t]{0.86\textwidth}
        \centering
        \textbf{Alignment-only}\par\smallskip
        \includegraphics[width=\linewidth]{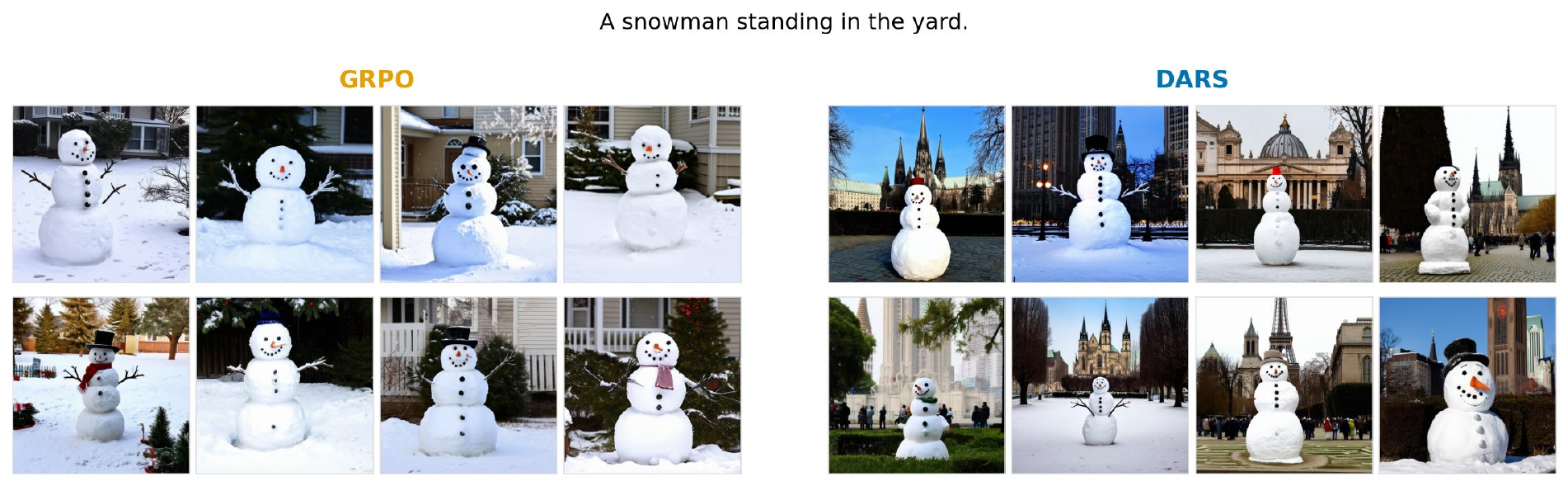}
    \end{minipage}

    \vspace{0.5em}

    \begin{minipage}[t]{0.86\textwidth}
        \centering
        \textbf{Combined}\par\smallskip
        \includegraphics[width=\linewidth]{Img_results/qwen_aes/prompt_001_rewrite_grid.pdf}
    \end{minipage}\par\vspace{0.4em}
    \begin{minipage}[t]{0.86\textwidth}
        \centering
        \textbf{Alignment-only}\par\smallskip
        \includegraphics[width=\linewidth]{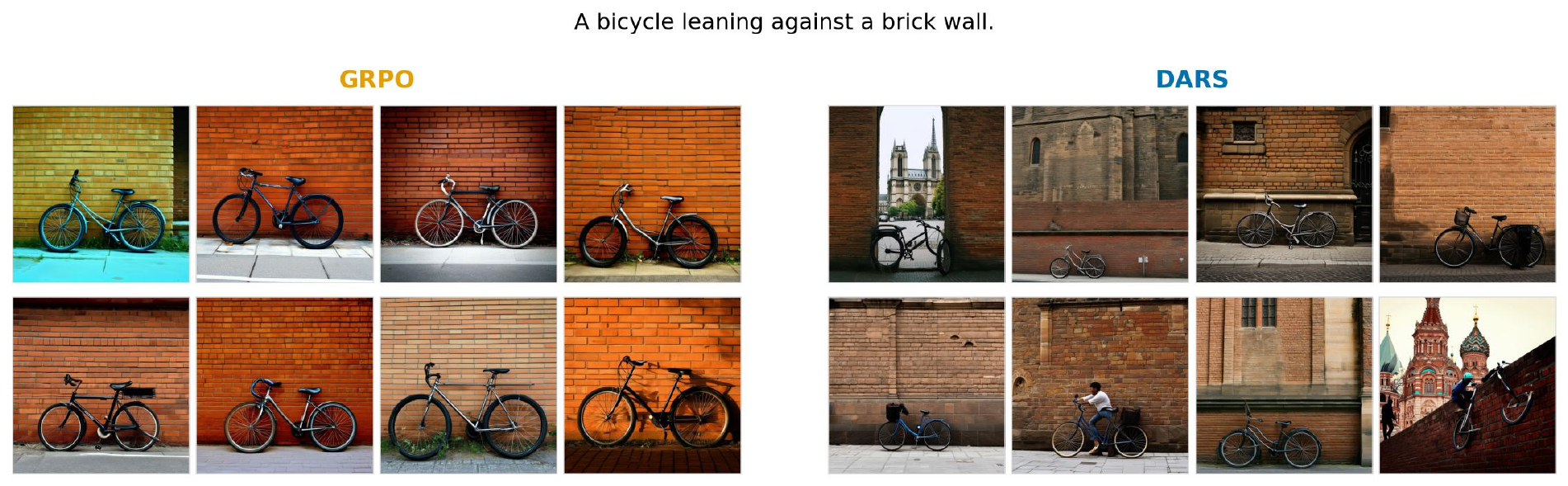}
    \end{minipage}
    \caption{Reward comparison for ``A snowman standing in the yard'' (top) and ``A bicycle leaning against a brick wall'' (bottom), using Qwen3-0.6B.}
    \label{fig:reward-comparison-01}
\end{figure*}

\begin{figure*}[t]
    \centering
    \begin{minipage}[t]{0.86\textwidth}
        \centering
        \textbf{Combined}\par\smallskip
        \includegraphics[width=\linewidth]{Img_results/qwen_aes/prompt_002_rewrite_grid.pdf}
    \end{minipage}\par\vspace{0.4em}
    \begin{minipage}[t]{0.86\textwidth}
        \centering
        \textbf{Alignment-only}\par\smallskip
        \includegraphics[width=\linewidth]{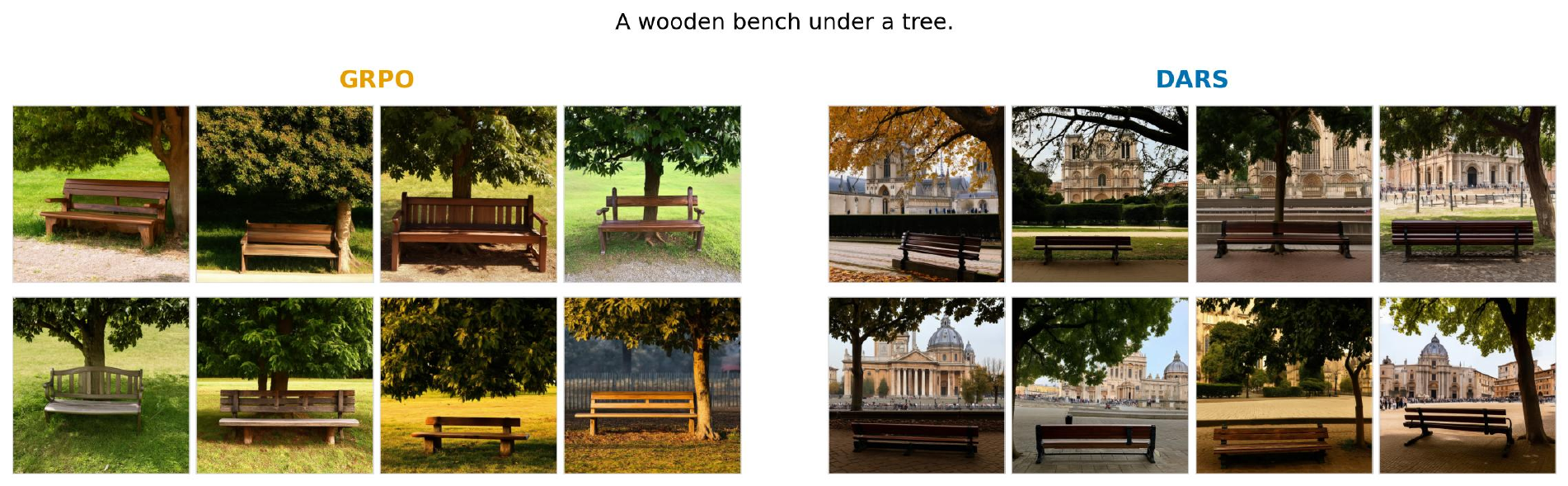}
    \end{minipage}

    \vspace{0.5em}

    \begin{minipage}[t]{0.86\textwidth}
        \centering
        \textbf{Combined}\par\smallskip
        \includegraphics[width=\linewidth]{Img_results/qwen_aes/prompt_003_rewrite_grid.pdf}
    \end{minipage}\par\vspace{0.4em}
    \begin{minipage}[t]{0.86\textwidth}
        \centering
        \textbf{Alignment-only}\par\smallskip
        \includegraphics[width=\linewidth]{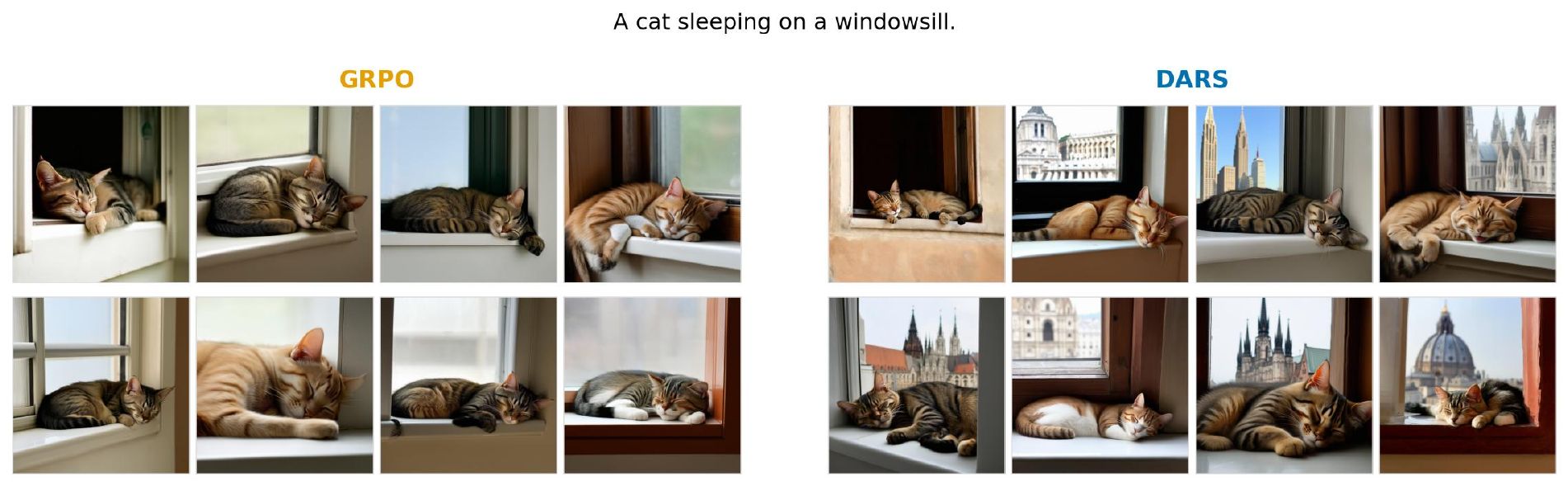}
    \end{minipage}
    \caption{Reward comparison for ``A wooden bench under a tree'' (top) and ``A cat sleeping on a windowsill'' (bottom).}
    \label{fig:reward-comparison-23}
\end{figure*}

\begin{figure*}[t]
    \centering
    \begin{minipage}[t]{0.86\textwidth}
        \centering
        \textbf{Combined}\par\smallskip
        \includegraphics[width=\linewidth]{Img_results/qwen_aes/prompt_004_rewrite_grid.pdf}
    \end{minipage}\par\vspace{0.4em}
    \begin{minipage}[t]{0.86\textwidth}
        \centering
        \textbf{Alignment-only}\par\smallskip
        \includegraphics[width=\linewidth]{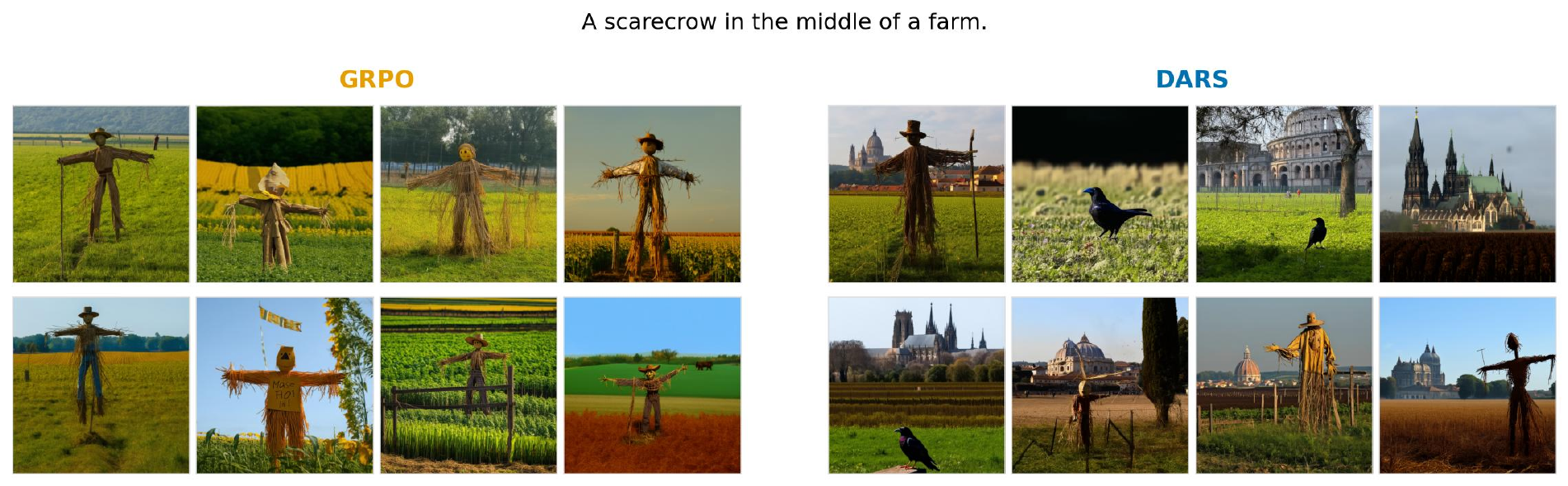}
    \end{minipage}

    \vspace{0.5em}

    \begin{minipage}[t]{0.86\textwidth}
        \centering
        \textbf{Combined}\par\smallskip
        \includegraphics[width=\linewidth]{Img_results/qwen_aes/prompt_005_rewrite_grid.pdf}
    \end{minipage}\par\vspace{0.4em}
    \begin{minipage}[t]{0.86\textwidth}
        \centering
        \textbf{Alignment-only}\par\smallskip
        \includegraphics[width=\linewidth]{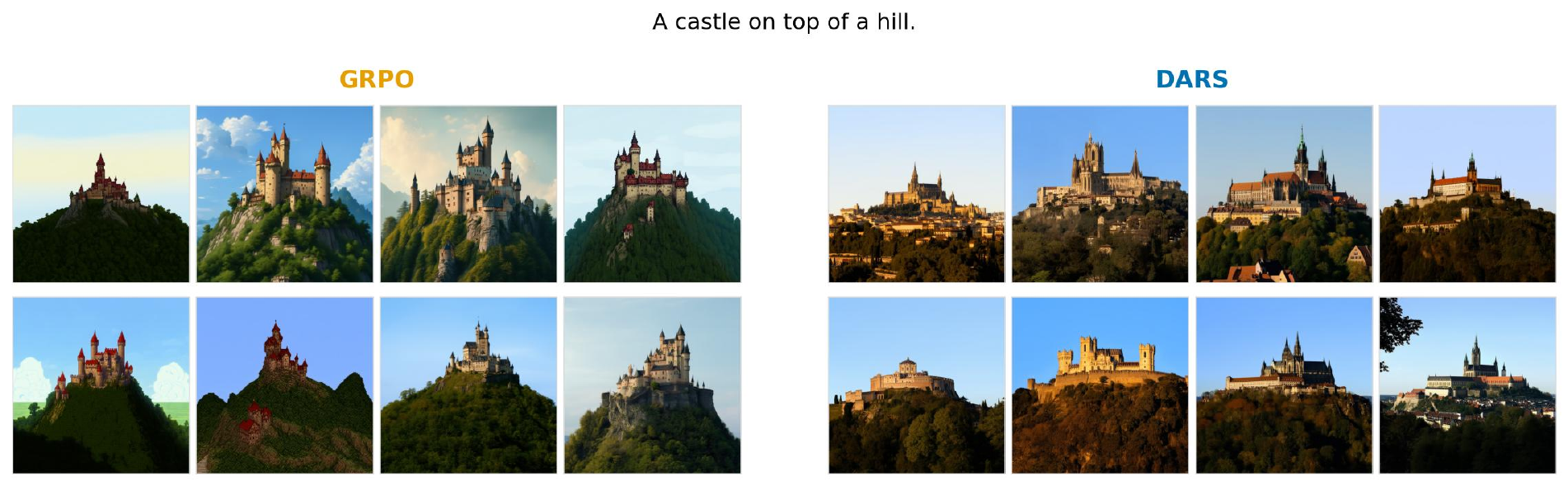}
    \end{minipage}
    \caption{Reward comparison for ``A scarecrow in the middle of a farm'' (top) and ``A castle on top of a hill'' (bottom).}
    \label{fig:reward-comparison-45}
\end{figure*}

\begin{figure*}[t]
    \centering
    \begin{minipage}[t]{0.86\textwidth}
        \centering
        \textbf{Combined}\par\smallskip
        \includegraphics[width=\linewidth]{Img_results/qwen_aes/prompt_006_rewrite_grid.pdf}
    \end{minipage}\par\vspace{0.4em}
    \begin{minipage}[t]{0.86\textwidth}
        \centering
        \textbf{Alignment-only}\par\smallskip
        \includegraphics[width=\linewidth]{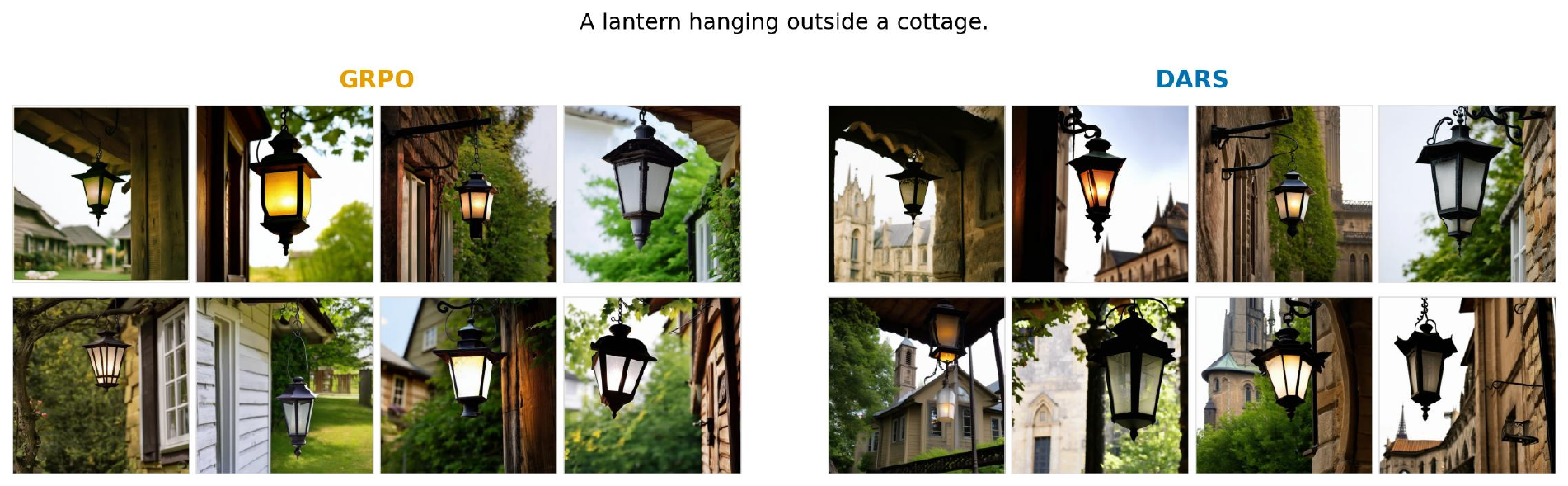}
    \end{minipage}

    \vspace{0.5em}

    \begin{minipage}[t]{0.86\textwidth}
        \centering
        \textbf{Combined}\par\smallskip
        \includegraphics[width=\linewidth]{Img_results/qwen_aes/prompt_007_rewrite_grid.pdf}
    \end{minipage}\par\vspace{0.4em}
    \begin{minipage}[t]{0.86\textwidth}
        \centering
        \textbf{Alignment-only}\par\smallskip
        \includegraphics[width=\linewidth]{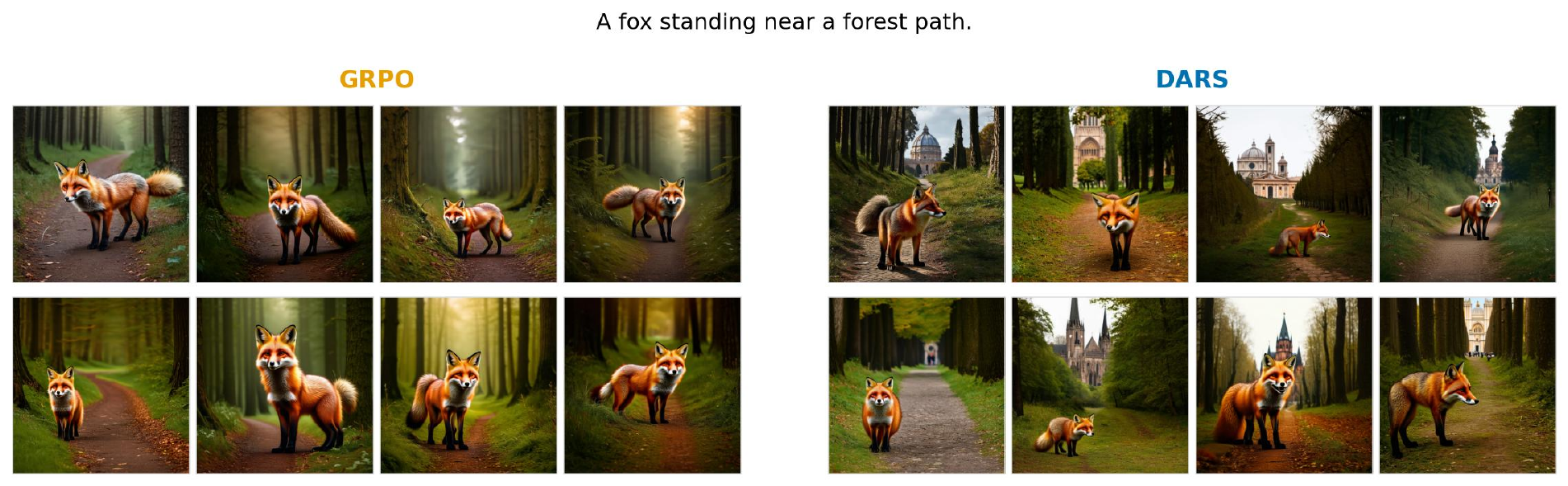}
    \end{minipage}
    \caption{Reward comparison for ``A lantern hanging outside a cottage'' (top) and ``A fox standing near a forest path'' (bottom). }
    \label{fig:reward-comparison-67}
\end{figure*}

Figures~\ref{fig:reward-comparison-01}--\ref{fig:reward-comparison-67} compare the two reward functions using the same Qwen3-0.6B policy LM. The \emph{combined} reward combines semantic fidelity with improvement under the frozen aesthetic scorer, whereas the \emph{alignment-only} reward applies only the minimum CLIP-alignment constraint. Each embedded grid compares GRPO and DARS under one reward, allowing the effect of reward design to be separated from the effect of density scaling.

The alignment-only objective permits DARS to explore a substantially broader visual region since it is free of aesthetic bounds that affect the combined reward. Across the examples, it produces larger changes in setting, illumination, camera distance, and composition. This is particularly visible for the bench, cat, scarecrow, and fox prompts in Figures \ref{fig:reward-comparison-23},\ref{fig:reward-comparison-45}, and \ref{fig:reward-comparison-67} where the generated scenes include a wider range of architectural and natural backgrounds. The quantitative results show the same pattern with higher Image Vendi Scores and DINO pairwise distance for alignment-only rewards.

\textbf{Additional qualitative comparisons}

\begin{figure*}[p]
    \centering
    \includegraphics[width=0.82\textwidth]{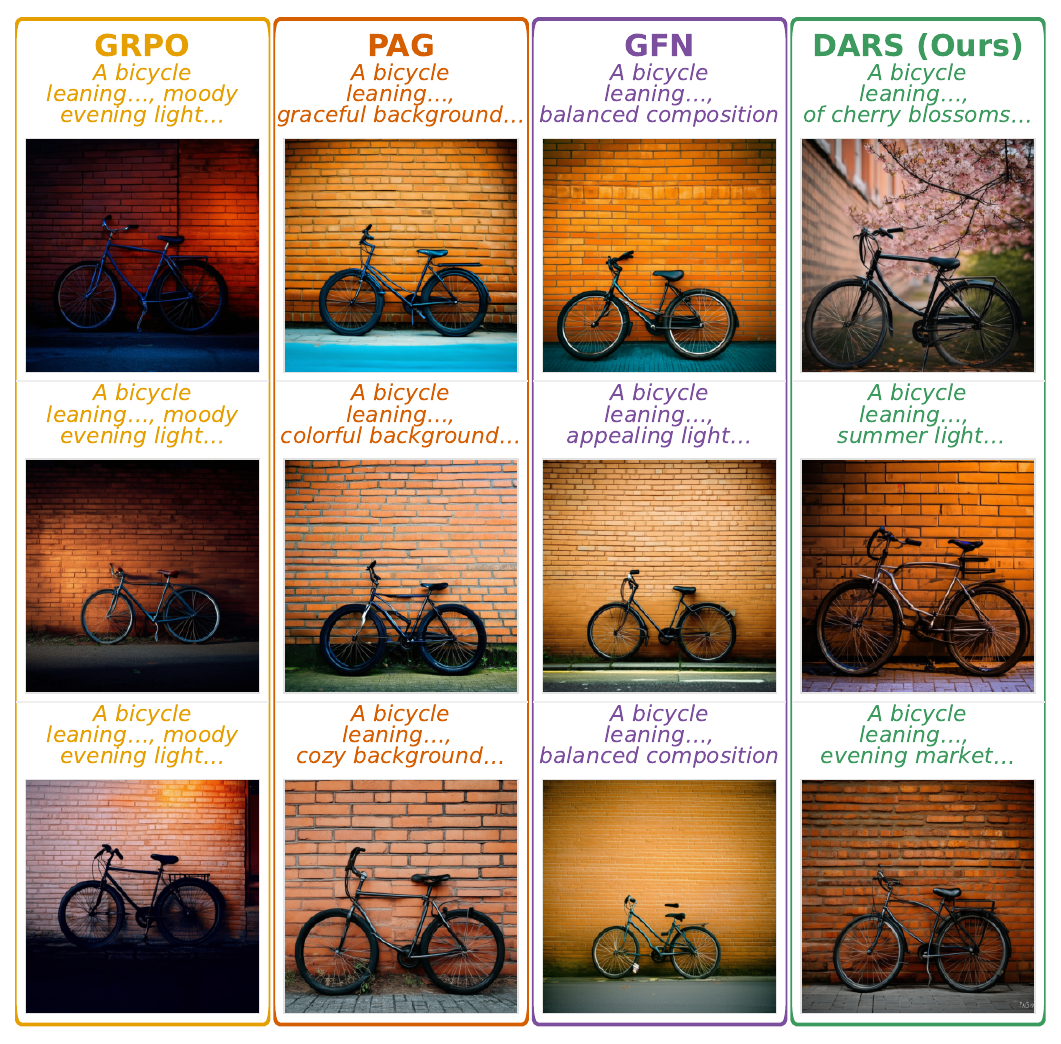}
    \caption{Qualitative comparison for ``A bicycle leaning against a brick wall.'' GRPO repeatedly generates a similar nighttime composition, while DARS-GRPO varies lighting, seasonal context, and surrounding details while preserving the bicycle and wall.}
    \label{fig:appendix-bicycle}
\end{figure*}

\begin{figure*}[p]
    \centering
    \includegraphics[width=0.82\textwidth]{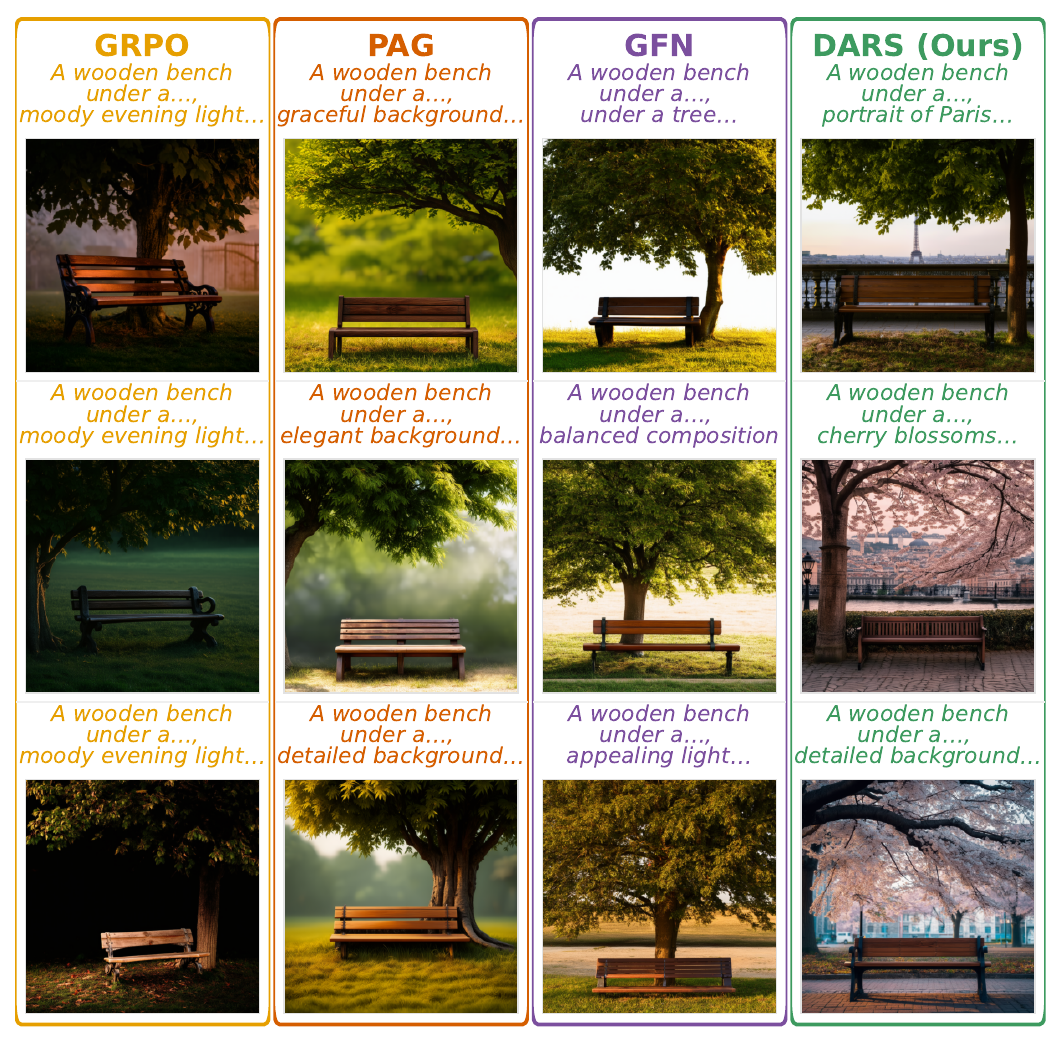}
    \caption{Qualitative comparison for ``A wooden bench under a tree.'' DARS-GRPO explores more varied locations, tree appearances, and visual contexts than GRPO, while PAG and GFN produce more limited changes around the same central composition.}
    \label{fig:appendix-bench}
\end{figure*}

\begin{figure*}[p]
    \centering
    \includegraphics[width=0.82\textwidth]{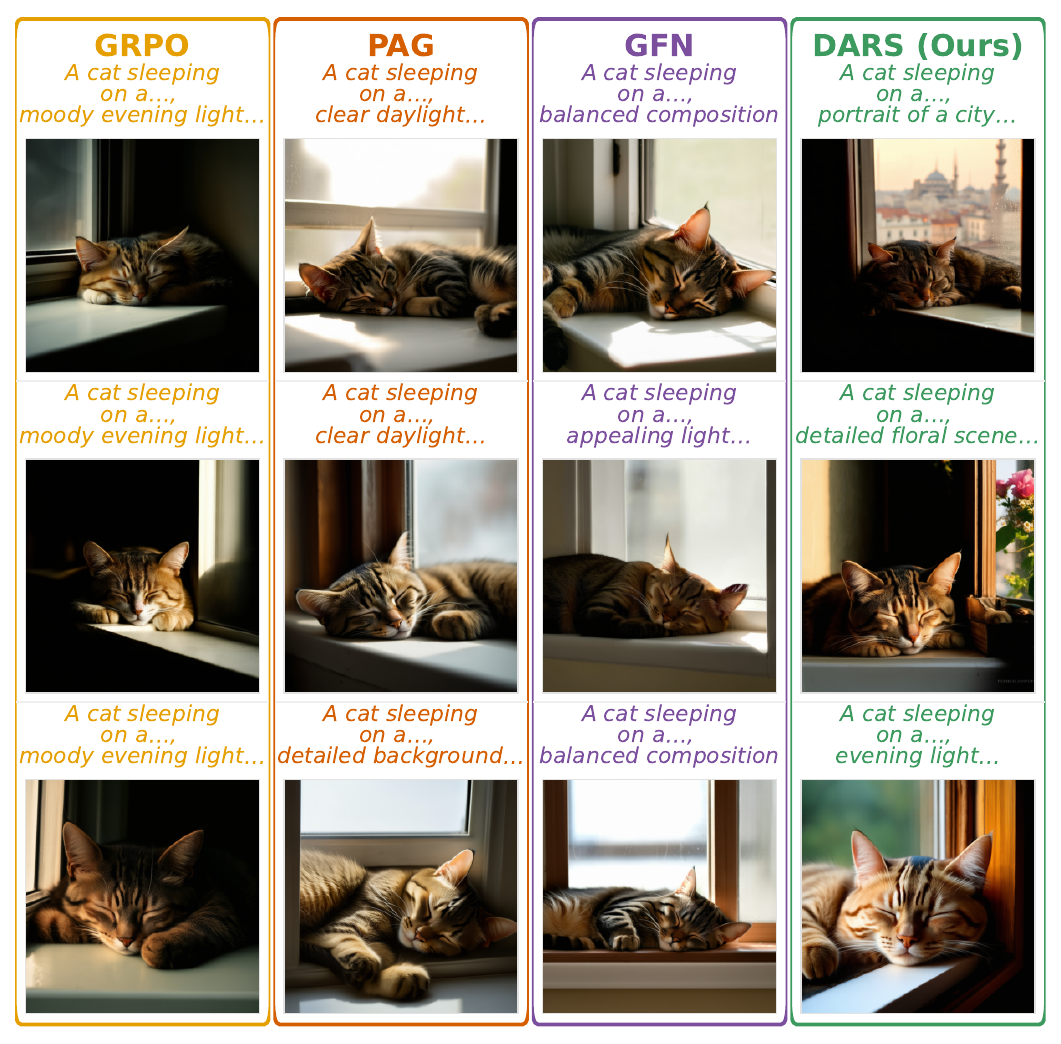}
    \caption{Qualitative comparison for ``A cat sleeping on a windowsill.'' All methods preserve the primary subject, while DARS-GRPO introduces broader variation in background scenery, lighting, and framing.}
    \label{fig:appendix-cat}
\end{figure*}

Figures~\ref{fig:appendix-bicycle}--\ref{fig:appendix-cat} provide additional
examples comparing GRPO, PAG, GFN, and DARS-GRPO. Across these prompts, GRPO
reuses a narrow rewriting pattern, whereas DARS produces more varied scene
contexts consequently leading to more diverse images.

\FloatBarrier

\subsection{Implementation Details}

\subsubsection{Experimental Configuration}

Unless otherwise specified, all experiments use the same training, generation, and evaluation settings. Table~\ref{tab:policy_config} summarizes the policy optimization hyperparameters shared by DARS and the GRPO baseline, while Table~\ref{tab:DARS_config} lists the hyperparameters specific to density-aware reward scaling. The generation pipeline, reward formulation, and datasets are held fixed across experiments unless explicitly varied in an ablation study.

\begin{table}[t]
\centering
\small
\caption{Policy optimization hyperparameters (DARS and GRPO).}
\label{tab:policy_config}
\resizebox{\columnwidth}{!}{%
\begin{tabular}{ll}
\toprule
\textbf{Parameter} & \textbf{Value} \\
\midrule
Policy LMs & SFT Qwen3-0.6B; SFT GPT-2 \\
Training temperature & 1.5 \\
Evaluation temperature & 1.0 \\
\texttt{max\_new\_tokens} & 64 \\
Learning rate & $1\times10^{-5}$ \\
Weight decay & 0.1 \\
Gradient clipping & 1.0 \\
PPO clipping $\epsilon$ & 0.25 \\
KL coefficient & $1\times10^{-9}$ \\
Policy updates per rollout & 1 \\
Training batch size & 2 prompts \\
\bottomrule
\end{tabular}}
\end{table}

DARS and GRPO are trained under identical optimization settings to ensure a fair comparison. Both methods use the same policy LM, optimizer hyperparameters, decoding configuration, and rollout schedule. They differ only in the reward scaling mechanism. Unless otherwise stated, these hyperparameters are used for all reported experiments.

\begin{table}[t]
\centering
\small
\caption{DARS-specific hyperparameters.}
\label{tab:DARS_config}
\resizebox{\columnwidth}{!}{%
\begin{tabular}{ll}
\toprule
\textbf{Parameter} & \textbf{Value} \\
\midrule
Density embedding space & \texttt{dinov2\_image} \\
Density kernel & Gaussian \\
Kernel width ($h$) & 20.0 \\

\bottomrule
\end{tabular}}
\end{table}
Table~\ref{tab:DARS_config} lists the hyperparameters specific to DARS. Density is estimated in the DINOv2 image embedding space using a Gaussian kernel with bandwidth $h=20$. To study the effect of density estimation, all kernel-width ablations retain the same training configuration while varying only the kernel bandwidth.

Kernel-width ablations use the same experimental setup while varying only the Gaussian kernel width,
\[
h \in \{10,15,20,25,30\},
\]
with all other hyperparameters kept fixed.

\subsubsection{Computing Environment}
All experiments were conducted on a server equipped with four NVIDIA A100-SXM4 GPUs using the software environment summarized in Table~\ref{tab:environment}. Mixed-precision training uses \texttt{bf16}, while Stable Diffusion inference is performed in \texttt{fp16} to improve computational efficiency without affecting evaluation.
\begin{table}[t]
\centering
\small
\caption{Hardware and software environment used for training.}
\label{tab:environment}
\resizebox{\columnwidth}{!}{%
\begin{tabular}{ll}
\toprule
\textbf{Component} & \textbf{Value} \\
\midrule
Operating System & Ubuntu 22.04.4 LTS (Linux 5.15.0-94-generic) \\
GPUs & $4\times$ NVIDIA A100-SXM4-40GB \\
Host RAM & $\sim$503 GiB \\
Python & 3.13.9 \\
PyTorch & 2.12.1+cu130 \\
Key Libraries & \texttt{transformers} 5.13.0, \texttt{diffusers} 0.39.0, \\
& \texttt{accelerate} 1.14.0, \texttt{peft} 0.19.1 \\
Mixed Precision & bf16 (\texttt{run.mixed\_precision}) \\
Diffusion Dtype & fp16 \\
\bottomrule
\end{tabular}%
}
\end{table}

\subsubsection{Training Configuration}

Tables~\ref{tab:policy_config} and~\ref{tab:DARS_config} summarize the optimization hyperparameters and DARS-specific settings used during training. Unless explicitly stated otherwise, all experiments employ these default configurations. DARS and GRPO differ only in the computation of policy rewards, while sharing identical optimization settings.

\subsubsection{Generation Configuration}
All generated images are produced using the Stable Diffusion v1.4 pipeline with identical sampling settings across methods. Each rewritten prompt is decoded independently using 20 diffusion steps with a classifier-free guidance scale of 7.5. The same CLIP encoder, DINOv2 encoder, and aesthetic predictor are used throughout both training and evaluation to ensure consistency across reported metrics.

\begin{table}[t]
\centering
\small
\caption{Shared generation and evaluation configuration.}
\label{tab:shared_config}
\resizebox{\columnwidth}{!}{%
\begin{tabular}{ll}
\toprule
\textbf{Parameter} & \textbf{Value} \\
\midrule
Diffusion model & CompVis/stable-diffusion-v1-4 \\
Resolution / Steps / CFG & $512\times512$, 20 steps, guidance scale 7.5 \\
Scheduler & DPM-Solver Multistep \\
\texttt{diffusion.seed\_mode} & \texttt{random} \\
\texttt{run.seed} & 42 \\
Rewrites per prompt ($G=N$) & 64 \\
CLIP encoder & OpenAI CLIP ViT-L/14 \\
DINOv2 encoder & \texttt{facebook/dinov2-base} (CLS embedding) \\
Aesthetic predictor & \texttt{sac+logos+ava1-l14-linearMSE} \\
CLIP alignment threshold & 0.20 \\
\bottomrule
\end{tabular}}
\end{table}

\subsubsection{Reward Configuration}
Two reward variants are considered in this work. The primary configuration combines aesthetic improvement with CLIP-based semantic alignment, while the alignment-only variant removes the aesthetic component to isolate the contribution of density-aware reward scaling. Apart from the reward definition, all experimental settings remain unchanged.

\begin{table}[t]
\centering
\small
\caption{Reward settings.}
\label{tab:reward_config}
\resizebox{\columnwidth}{!}{%
\begin{tabular}{ll}
\toprule
\textbf{Setting} & \textbf{Implementation} \\
\midrule
Combined &
Product-style aesthetic gain with CLIP alignment gating \\
Alignment only &
CLIP alignment reward without aesthetic gain \\
\bottomrule
\end{tabular}}
\end{table}

\subsubsection{Evaluation Protocol}

Evaluation follows the same protocol for all methods. For each evaluation prompt, the policy generates $N=64$ rewritten prompts, each of which is used to generate one image under identical diffusion sampling settings. Diversity metrics are computed only among images originating from the same source prompt and are then averaged across the evaluation set.

Image-text alignment is measured using OpenAI CLIP ViT-L/14, while image diversity is computed from DINOv2 image embeddings. The aesthetic LAION predictor (\texttt{sac+logos+ava1-l14-linearMSE}) is used wherever aesthetic reward is required.

During training, model checkpoints are periodically evaluated on the validation set using the same evaluation protocol. The checkpoint used for all reported experiments is selected based on the highest validation Vendi Score after reward saturation, ensuring that model selection aligns with the diversity objective optimized by DARS.

\subsubsection{Datasets}
The MS-COCO training split is used for policy optimization, while evaluation is performed on held-out MS-COCO prompts and the Lexica benchmark to assess both in-domain and out-of-domain generalization. Unless otherwise stated, all reported quantitative results use these dataset splits.
\begin{table}[t]
\centering
\small
\caption{Training and evaluation datasets.}
\label{tab:data_config}
\resizebox{\columnwidth}{!}{%
\begin{tabular}{ll}
\toprule
\textbf{Split} & \textbf{Dataset} \\
\midrule
Training & \texttt{dataset/coco/n\_1024/train.jsonl} \\
MS-COCO evaluation & \texttt{dataset/coco/n\_1024/test.jsonl} \\
Lexica evaluation & \texttt{dataset/lexica/n\_1024/test.jsonl} \\
\bottomrule
\end{tabular}}
\end{table}

\end{document}